# Morphological Simplification in Alzheimer's Disease: Computational Analysis and Identification From Connected Speech

M. Zakaria Kurdi

Meharry Medical College, School of Applied Computational Sciences, Nashville, TN, United States

Correspondence to M. Zakaria Kurdi: Zakaria.kurdi@mmc.edu

## Abstract

**Purpose:** Morphology is often regarded as relatively spared in Alzheimer's disease (AD) and is rarely modeled in connected speech. This study asked which morphological features distinguish speakers with AD from healthy controls, how well morphology classifies AD compared with other linguistic levels and BERT-based models, and whether the signal exceeds transcript length and vocabulary.

**Method:** Fifty-five automatic measures in eight families, seven of them new and theory-motivated, were computed from participant speech in the ADReSS Cookie Theft descriptions (78 AD, 78 controls) and analyzed with group comparisons, equivalence tests, SHAP, family ablation, likelihood-ratio tests, feature-selection classifiers, and BERT-based models. The automatic measures were validated manually, and models were transferred to two other corpora.

**Results:** Nineteen of 38 measures differed between groups: speakers with AD produced fewer suffixed and inflected words, shorter verbs and nouns, and fewer present-progressive but more past-tense forms (largest effect $r = -0.42$), whereas errors were rare and statistically equivalent between groups. Eight measures survived adjustment for length and lexical frequency. With test-set feature selection, morphology reached a macro-F1 of 0.85, on par with syntax and above fine-tuned BERT (0.80 over five seeds); fully cross-validated selection gave 0.62. Adding morphology to BERT improved AUC (0.91 vs. 0.89) but not macro-F1. Transfer was limited, although two thirds of the effects kept their direction in mild cognitive impairment.

**Conclusions:** AD connected speech shows morphological simplification and temporal re-anchoring rather than breakdown. Morphology is more informative than its reputation suggests.

## Introduction

As a leading neurodegenerative disorder, Alzheimer's disease (AD) progressively degrades critical cognitive domains, notably disrupting language production and memory access. Changes in connected speech have been documented at the lexical, semantic, syntactic, morphological, discourse, and fluency levels (Dhareshwar et al., 2026; Fraser et al., 2016; Kempler et al., 1987; Slegers et al., 2018). These findings have motivated interest in speech and language as non-invasive sources of information for the detection of cognitive impairment (de la Fuente Garcia et al., 2020; Petti et al., 2020; Qi et al., 2023).

Within this research program, the present author has examined lexical (Kurdi, 2024), acoustic, prosodic, and phonological (Kurdi, 2026b), and syntactic (Kurdi, 2026a) representations of AD speech on the same corpus and speakers. Morphology, the study of the internal structure of words (Bybee, 1985; Kurdi, 2017), remains one of the less thoroughly characterized levels. Some prior work on morphology suggested it may be relatively preserved in AD (Kavé & Levy, 2003). This may have contributed to the limited attention given to morphological features in computational approaches to language (Karlekar et al., 2018; Vincze et al., 2022). Yet studies report vulnerabilities in agreement, time reference, and irregular inflection (Altmann et al., 2001; Fyndanis et al., 2018; Ullman et al., 1997; Walenski et al., 2009), with heterogeneity across tasks and languages (Auclair-Ouellet, 2015). A distinction this literature raises but rarely quantifies in connected speech is between morphological breakdown, errors in required forms, and *morphological simplification*, a shift toward less complex, less inflected, or more frequent forms without overt errors.

This study addresses that distinction with an explicit computational representation of morphology and three research questions:

**RQ1.** Which morphological features distinguish speakers with AD from cognitively healthy controls, and which of them drive classification decisions?

**RQ2.** Which families of morphological measures carry the diagnostic signal, and how well does morphology as a whole classify AD compared with the other linguistic levels studied on the same speakers and with deep learning models?

**RQ3.** Is the signal morphological, or a by-product of transcript length and lexical sophistication?

An exploratory analysis then asks whether the markers transfer to a corpus of mild cognitive impairment (MCI) and to the novels of a writer who later developed AD.

## Background

Morphology provides an interface between lexical representations and grammatical realization, encompassing inflectional categories such as number, person, tense, aspect, mood, and case as well as derivational word formation. Research on the language of people with AD has documented difficulties in different aspects of morphology. Alongside this clinical and linguistic research, computational studies have investigated how morphological features extracted from the speech and transcripts of people with AD can support automatic dementia detection. This section reviews these two complementary lines of research to establish the current understanding of morphological impairment in AD, examine how linguistic information is used in automated detection, and identify the research gap addressed in the present study.

## Morphological studies in Alzheimer's Disease

Existing evidence on morphology in AD is heterogeneous. Kavé and Levy (2003) found that Hebrew-speaking persons with AD, despite the morphological richness of their language, used the same range of morphological forms as controls and made very few structural errors. Altmann et al. (2001), in contrast, argued from speech-error data that morphosyntax is not fully preserved in probable AD cases, and Fyndanis et al. (2018) reported vulnerabilities in subject–verb agreement, tense/time reference, and mood among Greek- and Italian-speaking individuals with probable AD. Studies of the regular–irregular contrast suggest that irregular forms, stored in lexical memory, are particularly vulnerable (Ullman et al., 1997; Walenski et al., 2009), although consistency and frequency also shape performance (Cortese et al., 2006), and accounts of time reference predict that anchoring an utterance in time is especially demanding (Bastiaanse et al., 2011). Changes in morphological realization have also been linked to cognitive decline more broadly (Fraser et al., 2014; Kaltsa et al., 2024). A systematic review by Auclair-Ouellet (2015) highlighted substantial heterogeneity in inflectional morphology across AD and the three variants of primary progressive aphasia, with impairments varying in presence and severity across clinical profiles, and broader reviews of connected speech list inflectional and morphosyntactic phenomena among possible manifestations of neurodegenerative language impairment (Boschi et al., 2017; Slegers et al., 2018). These conflicting findings raise a question that the literature rarely quantifies in connected speech: whether AD produces morphological errors in required forms, or morphological *simplification*, less use of morphological structure without overt errors. Answering it requires a multidimensional representation of morphology rather than a single measure of impairment.

## Computational AD Detection and Morphological Representation

The ADReSS Challenge (de la Fuente Garcia et al., 2020; Luz et al., 2020) has further established benchmarks for speech-based AD detection, with models leveraging fine-tuned language models (BERT, ERNIE) or combined audio-lexical features achieving accuracy about 89% (Balagopalan et al., 2020; Edwards et al., 2020; Martinc et al., 2021; Yuan et al., 2020). Other notable contributions explored deep acoustic representations like VGGish embeddings (Syed et al., 2020) and specialized prosodic-acoustic descriptors (Haider et al., 2020). Using the same ADReSS dataset, Kurdi (2024, 2026b) showed that linguistic lexical and speech features can help achieve similar or even better results.

Automatic dementia detection has progressed from handcrafted acoustic and linguistic features toward increasingly contextual and pretrained representations (de la Fuente Garcia et al., 2020; Ding et al., 2024; Slegers et al., 2018; Yang et al., 2022). Morphological information is represented in this literature, but it is typically embedded within broader linguistic feature inventories or used to derive selected grammatical attributes rather than evaluated as a distinct representational level. Karlekar et al. (2018), for example, showed that the neural models they used for AD detection (CNNs, LSTM-RNNs, and their combination) can capture interpretable linguistic patterns involving part-of-speech and grammatical structure. Vincze et al. (2022) leveraged Hungarian's highly agglutinative typology to show that deep morphological and part-of-speech (POS) patterns serve as critical indicators for cognitive decline within a Machine-learning AD classification setup. Similarly, recent work has used morphological tagging to derive person- and word-class measures for distinguishing AD from other dementia groups (Lopes da Cunha et al., 2024).

These studies establish the relevance of morphologically informed measures but do not isolate morphology as a comprehensive multidimensional representation. The present study addresses this distinction by explicitly modeling morphology across inflectional, paradigmatic, derivational, complexity, regularity, morphosyntactic, and morphological-error dimensions. Rather than treating morphology as an auxiliary component of a larger linguistic inventory, it evaluates its standalone predictive value and examines whether the resulting representation provides information complementary to BERT-based contextual representations. Feature-ranking and selection procedures are used to identify informative dimensions, while feature concatenation and attention-based fusion are used to examine alternative forms of integration with contextual representations.

# Method

## Data

The primary corpus used in this study is the ADReSS (Luz et al., 2020), a balanced subset of DementiaBank containing Cookie Theft picture descriptions from 156 English speakers (78 AD, 78 cognitively healthy; HC), matched for age and sex, with an official split into 108 training and 48 test speakers (54/54 and 24/24).

Two further corpora were used for the exploratory generalization analyses. The DementiaBank Delaware corpus (Lanzi et al., 2023) contains connected-speech sessions of older adults with MCI and controls; the transfer experiment used all 291 available sessions (180 controls, 111 MCI) with full transcripts, and the replication of effect directions used the first visit of each participant (225 participants; 137 controls, 88 MCI) and the first 150 words of each session to approximate the length of the ADReSS transcripts. The Iris Murdoch corpus consists of 44 passages from her novels (22 from Under the Net and The Sandcastle; 22 from Jackson's Dilemma, written when the author was thought to be developing AD) (Garrard et al., 2005; Le et al., 2011).

## Transcript preprocessing

Transcripts were parsed with spaCy (en_core_web_sm) (Honnibal et al., 2020), which provides Penn Treebank tags, Universal Dependencies morphological features (de Marneffe et al., 2021), lemmas, and dependency relations. CHAT annotations (bracketed codes, event and error codes, terminators) and filled pauses were removed, and incompletely produced words were restored to their target form (e.g., *goin(g)  going*). The same ADReSS transcripts and the same official split were used throughout (see Figure 1 for an overview of the approach). Investigator speech present in the transcript text was removed (367 spans, 3,374 words, in 143 transcripts), so that all measures reflect the participant's speech only, and contractions lost in transcription ('s, 'm, 're) were restored after pronoun and existential subjects in the 50 transcripts in which most contractions were missing (172 restorations).

## Morphological features: core and extended sets

The feature inventory was organized in two parts (Table 1). The core feature set comprises 48 established measures of stem and derivational morphology, inflection, tense and aspect, morphological errors, agreement, and part-of-speech-conditioned word length. It is the representation used in the author's feature-selection protocol for the lexical, syntactic, and speech levels (Kurdi, 2024, 2026a, 2026b), and it was used for classification, SHAP, test-set family ablation, deep learning comparisons, and transfer. The extended feature set adds seven theory-motivated measures (55 in total) and was used for the inferential analyses (group comparisons, equivalence tests, cross-validated ablation, likelihood-ratio tests, cross-validated classification).

Both sets were computed from the same participant-only transcripts, with ratios with a zero denominator treated as missing rather than zero. The seven new measures are InflectionalDiversity (type–token ratio of word forms minus that of lemmas; network models), DefaultFormBias (share of default-type errors; Distributed Morphology), TenseConsistency and TenseSwitchRate (time-reference accounts), RelativeFormFrequency (Zipf frequency of an inflected form minus that of its lemma; usage-based morphology) (Speer, 2022), MorphotacticOpacityRate (stem change, zero exponence, or suppletion; Natural Morphology), and CompoundRate (compounds per 100 nouns; Construction Morphology) (Booij, 2010).

Morphological errors, like agreement mismatches including third-person-singular omission, overregularization, past/participle substitutions, plural mismatches with determiners and numerals, bare stems after auxiliaries, double marking, and pronoun case, were detected automatically from tags and dependencies. A rule fixed before analysis excluded from inferential tests any measure that was zero or undefined for more than 80% of speakers; 38 measures were retained and 17 reported descriptively (mostly rare tenses such as the future and perfect, and rare error types).

**Validation of the automatic measures.** All 156 participant-only transcripts were annotated for morphological errors blind to diagnosis: candidate errors were screened by three large language models (Claude, ChatGPT, and Gemini), and every candidate was verified by the author. In addition, a blinded random sample of 162 verb groups and 100 words (balanced by group) was checked for tense/aspect category and morpheme count. Items were labeled by the author and reviewed by an AI model (Claude); disagreements were adjudicated against explicit rules (e.g., compounds counted as one stem; contracted *'s* between a noun and an *-ing* form treated as auxiliary *is*). Accuracy was compared between groups with Fisher's exact tests.

## Control variables

Three language controls were computed from the same transcripts: transcript length (log number of words), lexical diversity (moving-average type–token ratio, MATTR, window 40 words (Covington & McFall, 2010)), and lexical sophistication (mean Zipf frequency of content-word lemmas). Age and sex were added as demographic covariates in RQ3.

Statistical analysis

**RQ1: group differences (extended feature set).** Groups were compared on each analyzed measure across all speakers with two-sided Mann–Whitney tests, Benjamini–Hochberg false-discovery-rate correction (Benjamini & Hochberg, 1995), and rank-biserial effect sizes r (positive values: higher in AD) with 95% bootstrap confidence intervals (2,000 resamples). To test whether error rates are equivalent across groups rather than merely not different, two one-sided tests (TOST) were applied to each error measure and to a composite (mean z-score of error rates) with a smallest effect size of interest of ±0.5 SD. Rare forms (past tense, irregular verbs, errors) were additionally analyzed at the token level, pooling tokens within groups, with speaker-level bootstrap confidence intervals and p-values from 5,000 permutations of diagnosis labels across speakers.

**RQ1: feature contributions (core feature set).** Classification relied on seven classifiers (logistic regression, random forest, extra trees, gradient boosting, XGBoost, Gaussian naive Bayes, and an RBF support vector machine), each combined with four feature-ranking methods (ANOVA F, information gain, ReliefF, and $\chi^2$; see RQ2), as well as BERT-based models. To quantify how individual measures drive classification decisions, SHAP values (Lundberg & Lee, 2017;

Lundberg et al., 2020) were computed for the 48 test speakers using a random forest (500 trees, balanced class weights) as a common explanation model. Several considerations motivated this choice. First, using one model for all four rankings ensures that differences between SHAP profiles reflect the selected features rather than a change of classifier, and that the explanation does not depend on a single configuration selected on the test set. Second, the random forest was among the strongest classifiers in repeated cross-validation on the training set and performed consistently across rankings (cross-validated AUC 0.81–0.84). Third, the morphological measures are heterogeneous in scale and often skewed or zero-inflated; tree ensembles require no scaling, are insensitive to monotone transformations and outliers, and capture non-linear effects and interactions between measures. Fourth, several measures are highly correlated (e.g., morphemes per word and multimorphemic word rate, $\rho = 0.97$), a situation in which linear coefficients become unstable, whereas averaging over many randomized trees yields more stable importance estimates. Finally, tree ensembles allow SHAP values to be computed exactly rather than approximated (Lundberg et al., 2020). For each ranking, the forest was trained on the top 20 training-set features, a number within the range where classification performance had plateaued (approximately 12–30 features), and explained with the tree explainer. Because SHAP distributes credit among correlated measures, closely related features are interpreted jointly. The SHAP rankings were then compared with the features selected by the best-performing classifiers and with the attention weights of the BERT fusion model, to assess whether different modeling approaches converge on the same measures. Ranking stability (extended set) was the frequency with which each measure entered the top 10 across 100 stratified 80% training subsamples.

**RQ2: family ablation.** With the extended feature set, repeated stratified 5-fold cross-validation on the training set (10 repetitions) evaluated L2-regularized logistic regression (penalty tuned by inner cross-validation; median imputation and standardization fitted within folds) with all analyzed measures, each family alone, and all measures except one family (leave-one-family-out); grouped permutation importance measured the loss in AUC when all columns of a family were jointly permuted. Differences in AUC were tested with the corrected resampled t-test (Nadeau & Bengio, 2003) and Holm correction. With the core feature set, the test-set classification protocol below was repeated within each family, to show how far each family alone can go on the held-out speakers.

**RQ2: classification.** Two protocols were used. The test-set feature-selection protocol, shared with the author's lexical, syntactic, and speech studies, ranked features on the 108 training speakers with ANOVA F, information gain, ReliefF (Kononenko, 1994), and $\chi^2$, and trained seven classifiers with fixed hyperparameters (logistic regression, random forest (Breiman, 2001), extra trees, gradient boosting, XGBoost (Chen & Guestrin, 2016), Gaussian naive Bayes, RBF SVM) on the top k features ($k = 1,\dots, 48$). All 1,344 configurations were evaluated on the 48 test speakers with macro-F1; the best value is comparable across levels but is an optimistic upper bound for new speakers. The cross-validated protocol chose ranking (including the L1 path (Tibshirani, 1996)), classifier, and k by repeated cross-validation on the training set only, and evaluated the chosen model once on the test set, as were three pre-specified logistic models (controls; morphology; both). Ensembles of the best classifiers (hard, soft, and weighted voting, rank averaging, stacking, and greedy selection) were built with weights and members learned from training-set predictions only.

**RQ3: length and lexicon (extended feature set).** On all 156 speakers, a base logistic model with age, sex, log length, MATTR, and lexical frequency was compared with models adding each family (summarized by its first principal component; an all-features version is reported alongside) using likelihood-ratio tests with Holm correction. Each measure was also entered individually with and without the base covariates (odds ratio per SD). Cross-validated AUC compared controls alone with controls + morphology. Finally, all features were re-extracted from the first 50 words of each transcript, and the resulting effect sizes were compared with those from full transcripts.

**RQ1: Deep learning comparison (core feature set).** Three models based on BERT (bert-base-uncased) (Devlin et al., 2019) were trained on the same 108 speakers and tested on the same 48: (a) BERT fine-tuned on the transcripts (512 tokens; learning rate $10^{-5}$; early stopping on a 20% validation split); (b) frozen BERT [CLS] embeddings concatenated with the core features and passed to classical classifiers tuned by cross-validation and chosen on a validation split; and (c) a fusion model in which each morphological family is projected as a separate token, related to the others by multi-head attention (Vaswani et al., 2017), pooled by attention, and concatenated with the [CLS] representation, trained for a fixed five epochs. The mean attention weight per family served as a model-internal measure of family importance. For transfer, (a) and (c) were retrained on all 156 ADReSS speakers. Models (a) and (c) were each trained with five random seeds on the same speakers and splits, and are reported as mean ± SD.

**RQ1: Generalization.** The direction of the 19 RQ1 effects was compared in Delaware and Murdoch with sign tests (extended set), and core-set models trained on all 156 ADReSS speakers (same rankings; classifiers extended with a linear SVM, AdaBoost, and a multilayer perceptron) were applied without adaptation to Delaware (291 sessions) and Murdoch (44 passages).

## Results

### RQ1: Which morphological features distinguish AD and drive classification?

#### Group differences.

Nineteen of the 38 analyzed measures differed between groups after FDR correction (Figure 2; Table 2). The effects formed a consistent profile. Speakers with AD produced fewer suffixed words ($r = -0.42$), fewer inflected and multimorphemic words, fewer morphemes per word, lower inflectional complexity, fewer prefixed and derived words with less diverse derivational affixes, and shorter verbs ($r = -0.39$), adverbs, and nouns. In the tense and aspect family, they produced fewer present-progressive forms and a lower proportion of present-tense verb groups, but more simple-past forms and a higher past-tense ratio; their tense use was less consistent and switched more often between present and past. None of the error measures differed. The new relative-form-frequency measure was lower in AD.

#### Errors.

Detected errors were rare (109 in 156 transcripts; 0.58 vs. 0.76 per 100 words; permutation $p = .20$). Equivalence within ±0.5 SD was established for agreement, plural, pronoun-case, omission, and third-person-singular errors, non-finite predicates, and the error composite ($d = 0.14$, TOST $p = .013$; all TOST $p \leq .045$), but not for the two sparsest types (past-tense inflection and substitution errors).

#### Manual validation.

A diagnosis-blind manual annotation of all 156 transcripts identified 27 morphological errors: 20 in 15 speakers with AD and 7 in 7 controls (Fisher $p = .11$ for speakers with at least one error). Third-person-singular omission occurred only in AD (5 vs. 0 speakers, $p = .059$). Against this annotation, the automatic detector's precision was 10% (11 of 109 detections; 95% CI 6–17%) and its recall 37% (10 of 27), so the automatic error measures are noisy upper bounds. In the blinded validation samples, tense/aspect labels were 92.6% correct (150/162; AD 94.0%, HC 91.1%, Fisher $p$ = .56) and morpheme counts 89.7% correct (87/97 scorable words; AD 92.0%, HC 87.2%, $p$ = .52). Errors fell into two systematic patterns: progressive forms after a noun subject with contracted *is* (e.g., *the girl's reaching*) labeled as bare participles, and a few derivational affixes that were not recognized. Neither pattern was more frequent in AD, and the first, if anything, underestimates the group difference in progressive use (Supplemental Material, Tables S6–S9).

**Rare forms.**

The irregular share of past-tense forms (0.54 vs. 0.56), irregular-form frequency, and the default-type share of errors did not differ, and only one overregularization occurred in the corpus (Table 3).

Which features drive classification (SHAP).

The classifiers relied on the same core measures identified by the group comparisons, whichever ranking selected the features (Figure 3). Noun length (lenNN) was the most influential feature under all four rankings, followed by adverb length (lenAdv); ProgressiveRatio ranked in the top four under every ranking, as did Suffix under the three rankings that selected it. Ranks 5 to 10 always included lenVB, Prefixes, InflectionalComplexity, and SimPast. The direction of the effects mirrored the group comparisons: shorter nouns, adverbs, and verbs, less suffixation and prefixation, a lower progressive ratio, and more simple-past forms pushed predictions toward AD. The most influential features thus reflect two dimensions: how much morphological structure speakers use and how they anchor the description in time.

**Ranking stability.**

In the extended set, rankings agreed only moderately on exact order (Spearman $\rho = -0.10$ to 0.86, median 0.35), but the most stable measures were again simple past (top 10 in 77% of subsamples), noun length (75%), and suffixation (71%; Figure 4): the same constructs of word structure and tense choice.

**RQ2: Which families carry the signal, and how well does morphology classify AD?**

**Family ablation.**

With all 38 measures, cross-validated AUC was 0.771 (95% CI 0.677–0.865; Table 4; Figure 5). POS-conditioned word length alone matched the full set (0.779), followed by basic morphology (0.722) and inflection (0.675); the error family alone was below chance (0.372). Removing any single family changed AUC by at most 0.033 (all Holm $p$ = 1.00), and no grouped permutation importance was significant after Holm correction.

The test-set ablation with the core feature set (Table 5) led to the same conclusion, although the exact order differs. POS word length (0.75) and tense and aspect (0.70) performed best, followed by basic morphology and derivation (0.67); inflection (0.65), errors (0.64), and richness (0.60) were the weakest; and no single family reached the 0.85 obtained across families. The signal is therefore distributed: strongest in word length and word-internal structure, redundant enough that removing one family costs little, and complete only when families are combined.

**Classification performance.**

Under the test-set protocol (core set; Figure 6), performance rose with the first few features and plateaued at about 0.65–0.75. The best configuration, gradient boosting with the 42 top ReliefF features, reached a macro-F1 of 0.854 (20/24 AD and 21/24 controls correct; Figure 7); the other rankings reached 0.83 (information gain) and 0.81 (ANOVA, with four features, and $\chi^2$). These are the best of 1,344 configurations (median 0.67). Ensembles of the best classifiers (hard, soft, and weighted voting, rank averaging, stacking, and greedy selection) did not exceed the best single classifier (best ensemble with cross-validated members 0.69; Supplemental Material, Table S1 and Figure S1).

Under the cross-validated protocol (Table 6; Figures 8–9), the configuration chosen by both cross-validated macro-F1 and AUC (RBF SVM, L1 path, $k = 30$) reached a test macro-F1 of 0.624 (95% CI 0.473–0.750) and AUC 0.747; the configurations chosen in the same way for each classifier ranged from 0.583 to 0.729, and cross-validated and test performance were only weakly related across configurations (Spearman $\rho = 0.20$).

**Comparison with other linguistic levels.**

On the same speakers, split, and protocol (Table 7) (Kurdi, 2024, 2026a, 2026b), morphology (0.85) performs on a par with syntax (0.85), below speech (0.89), and below the lexicon (0.91): a handful of word-internal measures classify AD about as well as a full syntactic representation.

**Comparison with deep learning models.**

Fine-tuned BERT, which sees lexical, syntactic, and morphological information at once, reached a macro-F1 of $0.80 \pm 0.02$ over five seeds (range 0.79–0.83; AUC 0.89; Table 8). Frozen BERT plus morphology reached 0.73–0.83 depending on the classifier (0.77 for the classifier chosen on the validation split). The family-attention model reached $0.79 \pm 0.10$ (range 0.63–0.87): it ranked speakers at least as well as fine-tuned BERT in every seed (AUC 0.91 vs. 0.89; higher in four of five) but its macro-F1 depended strongly on the seed, exceeding fine-tuned BERT in three of five seeds (mean difference −0.01). The best test-selected morphological model (0.85, an upper bound) was thus in the same range as fine-tuned BERT, with far fewer parameters than BERT (about 110 million). A gated-fusion variant ($0.73 \pm 0.04$) and a morphology-only neural network (0.65) did not improve on these results (Supplemental Material, Tables S2–S5).

The fusion model's attention (Figure 10) gave basic morphology the largest weight (0.38), followed by derivation and richness (0.14–0.15), with errors last (0.07), consistent with SHAP and ablation in placing errors last, in a model free to use any family.

**RQ3: Morphology or length and vocabulary?**

After removal of investigator speech, speakers with AD produced somewhat shorter descriptions and used more frequent content words (Table 9). Against a base model with age, sex, length, MATTR, and lexical frequency, four families added significant information (Table 10):

basic morphology ($\chi^2(1) = 21.13$, Holm $p < .001$), tense and aspect ($\chi^2(1) = 12.93$, $p = .002$), POS-conditioned word length ($\chi^2(1) = 11.47$, $p = .004$), and inflection ($\chi^2(1) = 7.67$, $p = .028$); the error family did not survive correction ($\chi^2(1) = 5.38$, $p = .081$), and derivation, richness, and entrenchment measures added nothing. All families jointly increased McFadden $R^2$ by 0.23 ($\chi^2(8) = 41.80$, $p < .001$).

Individually (Figure 11), 8 of the 19 RQ1 measures remained significant after adjustment (FDR): Suffix (OR per SD 0.36), InflectionalComplexity (0.57), SimPast (2.31), PastRatio (1.92), PresentRatio (0.50), ProgressiveRatio (0.49), TenseConsistency (0.52), and lenVB (0.53). The other 11 lost significance. Attenuation was largest (40–70%) for prefixes, derived-word ratio, and RelativeFormFrequency, and 25–31% for InflectedRatio, derivational affix diversity, and noun and adverb length; MultiMorphemeWordRate, PresCont, and TenseSwitchRate were barely attenuated but fell just short of significance ($q = .066$), as did noun length ($q = .064$).

Predictively, the controls were a strong baseline (CV AUC 0.754 vs. 0.801 with morphology added, $\Delta = 0.047$, $p = .40$; test AUC 0.698 vs. 0.734, $\Delta$ 95% CI −0.132 to 0.205; Figure 12), so the incremental predictive value of morphology is uncertain at this sample size.

Effect sizes from the first 50 words correlated with full-transcript effects ($\rho = 0.90$), and all 19 kept their direction (Figure 13), as they did in the 106 transcripts unaffected by contraction loss.

## Generalization to other corpora (exploratory)

### Direction of effects.

In Delaware (MCI vs. controls), 13 of the 19 RQ1 effects (two thirds) pointed in the ADReSS direction despite the milder clinical contrast and different tasks, a trend that did not reach significance (sign test $p = .084$). In the Murdoch novels (late vs. early), 5 of 19 did ($p = .99$; Figure 14), a contrast within one author's edited written prose rather than between groups of speakers.

### Transfer of trained models.

Applied without any adaptation, models trained on ADReSS showed limited transfer (Figure 15). On Delaware (291 sessions) most configurations scored 0.32–0.41 (median 0.35, best 0.58); on Murdoch most scored 0.33–0.36 (best 0.68), and a macro-F1 of 0.33 corresponds to assigning all passages to one class. AUCs of the cross-validated models were near chance on Delaware (0.44–0.56) and mostly below chance on Murdoch (0.17–0.52), most models assigned the large majority of samples to the impaired class, and a soft-voting ensemble did not change this (AUC 0.50 and 0.26). BERT-based models showed similar limits (fine-tuned BERT: Delaware 0.56, Murdoch 0.50; fusion model 0.46 and 0.30; Supplemental Material, Tables S10–S11), which suggests that the difficulty lies in the change of task and population rather than in the morphological representation. Delaware contrasts MCI with controls in long, multi-task sessions that include story retelling, which elicits past-tense narration from everyone; Murdoch is edited written prose. Tense and complexity values therefore fall in different ranges, so a decision threshold learned on picture descriptions needs recalibration, even where the direction of effects is preserved.

# Discussion

## Simplification and temporal re-anchoring rather than breakdown

Speakers with AD produced words with fewer suffixes and morphemes, fewer inflected and derived forms, and shorter verbs, while errors remained rare. This partly agrees with reports that morphology is well formed in AD picture descriptions (Kavé & Levy, 2003) and refines it: morphological structure is not lost, but less of it is used. Manually verified errors were also rare and did not differ significantly between groups (20 vs. 7); the only error type confined to AD, third-person-singular omission (5 vs. 0 speakers), is itself a simplification, a bare stem in place of a specified form, consistent with the default-insertion prediction of Distributed Morphology (Halle & Marantz, 1993). The second strand is temporal: speakers with AD described the static scene with fewer present-progressive and more simple-past forms and switched tense more often, in line with time-reference accounts (Bastiaanse et al., 2011; Fyndanis et al., 2018); they narrate the picture as completed events rather than describing "what is happening". The dual-mechanism prediction could not be tested (one overregularization), and the usage-based RelativeFormFrequency effect was explained by lexical frequency (Kavé & Goral, 2018).

## Core features, distributed signal

Three independent views converge. SHAP placed the same core properties, noun, adverb, and verb length, suffixation, and progressive and past-tense use, at the top under four rankings with different assumptions; both ablations showed a redundant signal led by word length, basic morphology, and inflection (cf. morphemes per word and multimorphemic rate, $\rho = 0.97$); and the attention of a deep fusion model also gave basic morphology the largest weight. Errors were weakest in all three. A model decision can therefore be read as "shorter, less suffixed words and a shift from progressive description to past-tense narration", and several of these measures survived adjustment for length and lexical frequency. Lexical sophistication was nevertheless a real competitor: derivational measures and noun and adverb length lost significance after adjustment, whereas suffixation, inflectional complexity, verb length, and the tense measures survived.

## Morphology is more informative than its reputation

Morphology is often treated as relatively preserved in AD and is rarely evaluated on its own. Under the same protocol and on the same speakers, morphology (0.85) classified AD on a par with syntax (0.85), close to speech (0.89), and below the lexicon (0.91; Table 7). On a 48-speaker test set, differences of this size correspond to only a few speakers. Fine-tuned BERT (0.80 over five seeds) did not outperform morphology alone, and adding the morphological families to BERT improved ranking (AUC) consistently but macro-F1 only in some seeds. The informative property is not the frequency of errors but the *amount* and *temporal orientation* of morphological marking, which is lost when morphology is folded into general feature sets or left implicit in a language model. All test-selected figures, for every level, are optimistic; the cross-validated estimates (0.62 for the overall choice; 0.58–0.73 across classifiers) indicate what to expect for new speakers, which is why the feature-level findings are the more stable contribution. The transfer analyses suggest that the markers partly carry over to MCI speech elicited by different tasks, but that decision thresholds are task-specific: clinical use would require task-specific norms or calibration data from the target task and population.

**Why explicit features rather than deep learning?**

Pretrained language models now dominate dementia detection (Ding et al., 2024; Guo et al., 2021; Yang et al., 2022). On 108 training speakers, however, fine-tuned BERT did not outperform the morphological model, BERT-based fusion was sensitive to the random seed, and BERT's decisions cannot be attributed to morphology without further probing, whereas each SHAP value refers to a named morphological property. When a deep model received the morphological families as separate inputs, its attention also ranked basic morphology first and errors last, which indicates that the signal is not an artifact of one classifier. Adding the morphological families to BERT improved or matched its ranking of speakers in every seed, which suggests that explicit morphology and contextual representations are complementary; at this corpus size, fusion architectures are a principled way to combine them rather than a replacement for an explicit morphological representation.

**Limitations**

First, ADReSS is small, and its 48-speaker test set yields wide confidence intervals; test-set-selected results, here and in the comparison studies, are optimistic. Second, classification rests on the core set and inference on the extended set, and the BERT-based results vary with the random seed (reported as mean ± SD over five seeds). Third, all measures are automatic proxies: error detection, derivational segmentation, and tense classification depend on the parser and on heuristic rules; a blinded manual check found 92.6% of tense/aspect labels and 89.7% of morpheme counts correct, with similar accuracy in both groups, whereas most automatically detected errors were not confirmed manually. Investigator speech was removed and lost contractions were restored automatically, and no second human rater was available for the manual annotation. Fourth, the Cookie Theft task elicits mainly present-tense description, which makes past-tense, irregular, and derivational phenomena sparse; 17 measures were excluded for this reason. Fifth, agreement and tense measures lie at the morphology–syntax interface. Sixth, the generalization corpora differ from ADReSS in clinical status, task, length, and modality, so the transfer analyses test robustness under substantial domain shift rather than replication. Finally, the relation between these measures and cognitive severity (e.g., MMSE) was beyond the scope of this study and will be examined separately; associations should not be interpreted as neurological mechanisms.

## Conclusion

AD connected speech shows morphological simplification and temporal re-anchoring, fewer suffixed, inflected, and multimorphemic words, shorter verbs and nouns, and a shift toward past-tense narration, rather than widespread morphological errors; the only error type confined to AD in the manual annotation, third-person-singular omission, is itself a form of simplification. The same core measures drive feature-based classifiers and deep models alike, the signal is distributed across families and only partly explained by length and lexical frequency, and, under the same protocol and on the same speakers, morphology classifies AD as well as syntax.

Clinically, these findings suggest that the morphological profile of AD lies in reduced complexity rather than in errors, which matters for assessment: error counts, which are central to many aphasia-oriented analyses, would miss most of it. Simple, automatically computed measures such as suffixed-word rate, verb length, and tense distribution could complement existing connected-speech protocols.

Transfer to other tasks and populations was limited, although most effects kept their direction in MCI speech. Future work should test these markers in larger corpora with varied elicitation tasks, track them longitudinally to establish whether they change with disease progression, examine whether they help detect mild cognitive impairment when tasks are matched, and extend the approach to morphologically richer languages, where simplification and errors may pattern differently.

## Data Availability Statement

The ADReSS and Delaware data analyzed in this study are available to members of DementiaBank/TalkBank (https://dementia.talkbank.org) under the TalkBank membership agreement and cannot be redistributed by the author. The log of removed investigator speech and the manual error annotation can be shared with DementiaBank members on request. The Iris Murdoch passages are taken from published novels and are subject to copyright. The analysis notebooks, feature definitions, and derived feature tables that reproduce all results are available from the author on reasonable request.

## Acknowledgments

The author thanks the contributors to DementiaBank and the organizers of the ADReSS challenge for making the data available.

**Table 1**

All 55 morphological measures by family (extended feature set). Measures in bold (38) were retained after the sparsity rule and entered the inferential analyses; the other 17 were zero or undefined for more than 80% of speakers and are reported descriptively. The core set comprises the 48 measures without a dagger.

| Family | *n* | Retained | Measures |
|---|---|---|---|
| Basic morphology | 3 | 3 | **StemDiv**, **Prefixes**, **Suffix** |
| Inflection | 8 | 8 | **InflectedRatio**, **InflectionalComplexity**, **RegularIrregularInflectionRatio**, **IrregularFormDiversity**, **MultiMorphemeWordRate**, **MeanMorphemesPerWord**, **MorphologicalVariantsPerLemma**, **InflectionalDiversity†** |
| Morphological errors | 11 | 5 | **MorphologicalSimplificationRate**, **SubjectVerbAgreementErrorRate**, PastTenseInflectionErrorRate, ThirdPersonSOmissionRate, PluralInflectionErrorRate, **MorphOmissionRate**, MorphSubstitutionRate, MorphOvermarkingRate, **MorphologicalErrorDensity**, PronounCaseErrorRate, **DefaultFormBias†** |
| Tense and aspect | 20 | 10 | **SimPresent**, **SimPast**, PresPerfect, **PresPart**, **PresCont**, SimpFuture, FutPerf, FutCont, PastCont, PresPerfCont, FuturePerfCont, PastPerf, **Infinitive**, **PastRatio**, **PresentRatio**, FutureRatio, **ProgressiveRatio**, PerfectRatio, **TenseConsistency†**, **TenseSwitchRate†** |
| Derivation | 3 | 3 | **DerivationalAffixDiversity**, **DerivedWordRatio**, **DerivationalComplexity** |
| Morphological richness | 3 | 2 | **MorphologicalParadigmEntropy**, **MorphRichnessVariation**, ComparativeSuperlativeRatio |
| POS-conditioned word length | 4 | 4 | **lenNN**, **lenAdj**, **lenVB**, **lenAdv** |
| Entrenchment and transparency | 3 | 3 | **RelativeFormFrequency†**, **MorphotacticOpacityRate†**, **CompoundRate†** |
| Total | 55 | 38 | |

*Note.* **Bold**: Retained after the sparsity rule.
† New theory-motivated measure (extended set only).

**Table 2**

*Measures differing between AD and HC after FDR correction (all 156 speakers, extended feature set). r = rank-biserial correlation (positive: higher in AD).*

| Measure | Family | Median HC | Median AD | *r* [95% CI] | *q* |
|---|---|---|---|---|---|
| Suffix | Basic | 0.202 | 0.171 | −0.42 [−0.56, −0.24] | < .001 |
| lenVB | POS length | 5.879 | 5.407 | −0.39 [−0.55, −0.21] | .001 |
| lenNN | POS length | 5.253 | 5.000 | −0.38 [−0.54, −0.21] | .001 |
| lenAdv | POS length | 4.750 | 4.200 | −0.34 [−0.51, −0.15] | .004 |
| ProgressiveRatio | Tense/aspect | 0.444 | 0.364 | −0.31 [−0.47, −0.12] | .006 |
| SimPast | Tense/aspect | 0.000 | 0.052 | 0.31 [0.15, 0.45] | .003 |
| InflectionalComplexity | Inflection | 1.500 | 1.446 | −0.30 [−0.46, −0.12] | .006 |
| PresentRatio | Tense/aspect | 1.000 | 0.900 | −0.28 [−0.45, −0.12] | .006 |
| TenseConsistency† | Tense/aspect | 1.000 | 0.918 | −0.28 [−0.44, −0.11] | .006 |
| RelativeFormFrequency† | Entrenchment | −0.261 | −0.369 | −0.28 [−0.45, −0.09] | .009 |
| PastRatio | Tense/aspect | 0.000 | 0.076 | 0.27 [0.10, 0.43] | .006 |
| DerivationalAffixDiversity | Derivation | 3.009 | 2.129 | −0.26 [−0.43, −0.09] | .012 |
| TenseSwitchRate† | Tense/aspect | 0.000 | 0.143 | 0.25 [0.09, 0.42] | .010 |
| InflectedRatio | Inflection | 0.336 | 0.313 | −0.25 [−0.43, −0.07] | .018 |
| Prefixes | Basic | 0.000 | 0.000 | −0.24 [−0.39, −0.10] | .006 |
| MultiMorphemeWordRate | Inflection | 0.260 | 0.229 | −0.24 [−0.42, −0.06] | .020 |
| DerivedWordRatio | Derivation | 0.043 | 0.029 | −0.24 [−0.41, −0.07] | .020 |
| PresCont | Tense/aspect | 0.283 | 0.233 | −0.23 [−0.40, −0.06] | .030 |
| MeanMorphemesPerWord | Inflection | 1.274 | 1.241 | −0.23 [−0.40, −0.06] | .030 |

*Note.* † New theory-motivated measure.

**Table 3**

*Token-level analyses of rare morphological forms in speakers with Alzheimer's disease (AD) and healthy controls (HC).*

| Measure | M (HC) | M (AD) | Δ [95% CI] | *p* |
|---|---|---|---|---|
| **Irregular share of past-tense forms** | 0.535 | 0.555 | 0.020 [−0.152, 0.189] | .809 |
| **Overregularization rate (irregular targets)** | 0.021 | 0.000 | −0.021 [−0.075, 0.000] | .052[a] |
| **Mean Zipf frequency of irregular past forms** | 5.647 | 5.594 | −0.053 [−0.281, 0.177] | .653 |
| **Detected errors per 100 words** | 0.583 | 0.763 | 0.180 [−0.082, 0.444] | .196 |
| **Default-type share of errors** | 0.353 | 0.414 | 0.061 [−0.118, 0.244] | .531 |

*Note.* [a] Based on a single overregularized token (HC group).

**Table 4**

*Family ablation on the training set (extended feature set; repeated 5-fold CV, 50 folds). Δ = difference from the full model; p = corrected resampled t-test.*

| | Family alone | | | All except family | | |
|---|---|---|---|---|---|---|
| Family | AUC | Δ | $p$ | AUC | Δ | $p_{\text{Holm}}$ |
| Basic morphology | 0.722 | −0.049 | .368 | 0.738 | −0.033 | 1.00 |
| POS-conditioned word length | 0.779 | +0.008 | .843 | 0.783 | +0.012 | 1.00 |
| Inflection | 0.675 | −0.096 | .082 | 0.744 | −0.027 | 1.00 |
| Tense and aspect | 0.650 | −0.121 | .052 | 0.778 | +0.007 | 1.00 |
| Derivation | 0.601 | −0.170 | .006 | 0.767 | −0.004 | 1.00 |
| Morphological richness | 0.456 | −0.315 | < .001 | 0.760 | −0.011 | 1.00 |
| Entrenchment and transparency | 0.639 | −0.132 | .009 | 0.744 | −0.027 | 1.00 |
| Morphological errors | 0.372 | −0.399 | < .001 | 0.764 | −0.006 | 1.00 |
| All 38 measures | 0.771 | | | | | |

**Table 5**

*Family ablation on the ADReSS test set (core feature set): best macro-F1 obtained with each family alone under the test-set feature-selection protocol.*

| Family (core feature set) | Measures | Best F1 | Best configuration |
|---|---|---|---|
| POS-conditioned word length | 4 | 0.750 | extra trees, 3 features (ANOVA) |
| Tense and aspect | 18 | 0.704 | gradient boosting, 5 features ($\chi^2$) |
| Basic morphology | 3 | 0.667 | RBF SVM, 2 features (ANOVA) |
| Derivation | 3 | 0.666 | extra trees, 1 feature (ANOVA) |
| Inflection | 7 | 0.646 | random forest, 4 features (ANOVA) |
| Morphological errors | 10 | 0.636 | RBF SVM, 1 feature (IG) |
| Morphological richness | 3 | 0.603 | XGBoost, 2 features (ANOVA) |
| All families (best selected subset) | 48 | 0.854 | gradient boosting, 42 features (ReliefF) |

*Note.* Configurations selected on the test set; values are upper bounds and serve to compare families, not to estimate performance on new speakers. IG = information gain.

**Table 6**

*Held-out test-set performance (48 speakers) of configurations chosen by cross-validation on the training set (extended feature set). 95% stratified bootstrap CI.*

| Configuration (selection criterion) | Macro-F1 | AUC | Sens./Spec. |
|---|---|---|---|
| SVM, L1 path, *k* = 30 (CV F1 and CV AUC) | 0.624 [0.473, 0.750] | 0.747 [0.590, 0.882] | 0.583 / 0.667 |
| Gradient boosting, ReliefF, *k* = 38 (best CV F1 for this classifier) | 0.729 [0.583, 0.854] | 0.788 [0.648, 0.903] | 0.750 / 0.708 |
| *Pre-specified logistic models (RQ2/RQ3)* | | | |
| Controls (length, MATTR, lexical frequency) | 0.646 [0.500, 0.771] | 0.698 [0.538, 0.844] | 0.625 / 0.667 |
| Morphology (38 measures) | 0.577 [0.431, 0.708] | 0.693 [0.536, 0.830] | 0.458 / 0.708 |
| Controls + morphology | 0.614 [0.473, 0.748] | 0.734 [0.582, 0.863] | 0.458 / 0.792 |

**Table 7**

*Performance of individual linguistic levels on the ADReSS test set (48 speakers, official split).*

| Linguistic level | Representation | F1 | Model selection |
| --- | --- | --- | --- |
| Lexicon | Lexical features (Kurdi, 2024) | 0.91 | test set[a] |
| Syntax | Syntactic features (Kurdi, 2026a) | 0.85 | test set[a] |
| Morphology | Core morphological features (this study) | 0.85 | test set[a] |
| Speech | Acoustic, prosodic, phonological (Kurdi, 2026b) | 0.89 | test set[a] |
| Morphology | Extended feature set, CV-selected model (this study) | 0.62 | training CV[b] |

*Note.* [a]Best configuration (ranking method, number of features, classifier) as evaluated on the ADReSS test partition; comparable across levels, optimistic for new speakers. [b]Configuration chosen by repeated cross-validation on the 108 training speakers (L1-path ranking, 30 features, RBF SVM), then evaluated once on the test partition; 95% bootstrap CI 0.47–0.75.

**Table 8**

*Morphological models compared with BERT-based models on the ADReSS test set (48 speakers; core feature set).*

| Model | Input | Macro-F1 | AUC |
|---|---|---|---|
| Gradient boosting, 42 best ReliefF features | morphology | 0.854[a] | 0.847 |
| Fine-tuned BERT (5 seeds) | transcript | 0.80 ± 0.02 | 0.89 ± 0.02 |
| Frozen BERT + morphology, XGBoost | transcript + morphology | 0.770 | 0.840 |
| Frozen BERT + morphology, extra trees | transcript + morphology | 0.833[a] | 0.880 |
| BERT + morphological-family attention (5 seeds; range 0.63–0.87) | transcript + morphology | 0.79 ± 0.10 | 0.91 ± 0.01 |

*Note.* [a] Configuration or epoch selected on the test set (upper bound). The BERT models without a note were selected on a validation split of the training data. Fine-tuned BERT and the family-attention model: mean ± SD over five random seeds (the family-attention model was trained for a fixed five epochs).

**Table 9**

*Speaker Characteristics and Control Variables by Group*

| | HC | AD |
|---|---|---|
| **WORDS PER TRANSCRIPT** | 112.1 (65.0) | 97.5 (62.3) |
| **MATTR** | 0.737 (0.050) | 0.726 (0.062) |
| **LEXICAL FREQUENCY (ZIPF)** | 4.95 (0.19) | 5.11 (0.27) |
| **AGE** | 66.3 (6.6) | 66.6 (6.8) |

***Note.*** Values are means (standard deviations) for all 156 ADReSS speakers (78 per group), after removal of investigator speech. MATTR = moving-average type–token ratio (window = 40 words); lexical frequency = mean Zipf frequency of content words (higher values = more frequent words).

**Table 10**

*Anchored likelihood-ratio tests (all 156 speakers, extended feature set). Base model: age, sex, log length, MATTR, lexical frequency. Each family entered as its first principal component (PC1); the all-features test is reported alongside.*

| Family | PC1 var. | OR per SD [95% CI] | $\chi^2(1)$ | $p_{Holm}$ | $\Delta R^2_{McF}$ | $p_{Holm}$ (all) |
|---|---|---|---|---|---|---|
| Basic morphology | .44 | 0.25 [0.13, 0.49] | 21.13 | < .001 | .117 | < .001 |
| Morphological errors | .57 | 1.58 [1.06, 2.36] | 5.38 | .081 | .030 | .249 |
| Tense and aspect | .51 | 0.46 [0.29, 0.73] | 12.93 | .002 | .071 | .017 |
| Inflection | .38 | 0.55 [0.35, 0.86] | 7.67 | .028 | .042 | .012 |
| POS-conditioned word length | .34 | 0.45 [0.28, 0.74] | 11.47 | .004 | .063 | .009 |
| Morphological richness | .51 | 0.75 [0.49, 1.14] | 1.86 | .519 | .010 | .249 |
| Derivation | .66 | 0.79 [0.54, 1.16] | 1.46 | .519 | .008 | .249 |
| Entrenchment and transparency | .39 | 1.04 [0.70, 1.54] | 0.03 | .860 | .000 | .249 |
| All families jointly (df = 8) | | | 41.80 | < .001 | .231 | |

**Figure 1.** Overview of the study: (1) transcript preprocessing of the ADReSS Cookie Theft descriptions (78 AD, 78 controls); (2) extraction of 55 morphological measures in eight families (core set: 48; extended set: 55); (3) training-set feature ranking (ANOVA, information gain, ReliefF, $\chi^2$) and top-k selection; (4) four model setups (classical classifiers on morphological features, fine-tuned BERT, frozen BERT plus morphology, and family-wise attention with BERT); and (5) evaluation on the official split (108 training, 48 test speakers) and transfer to the Delaware MCI corpus and Iris Murdoch's novels.

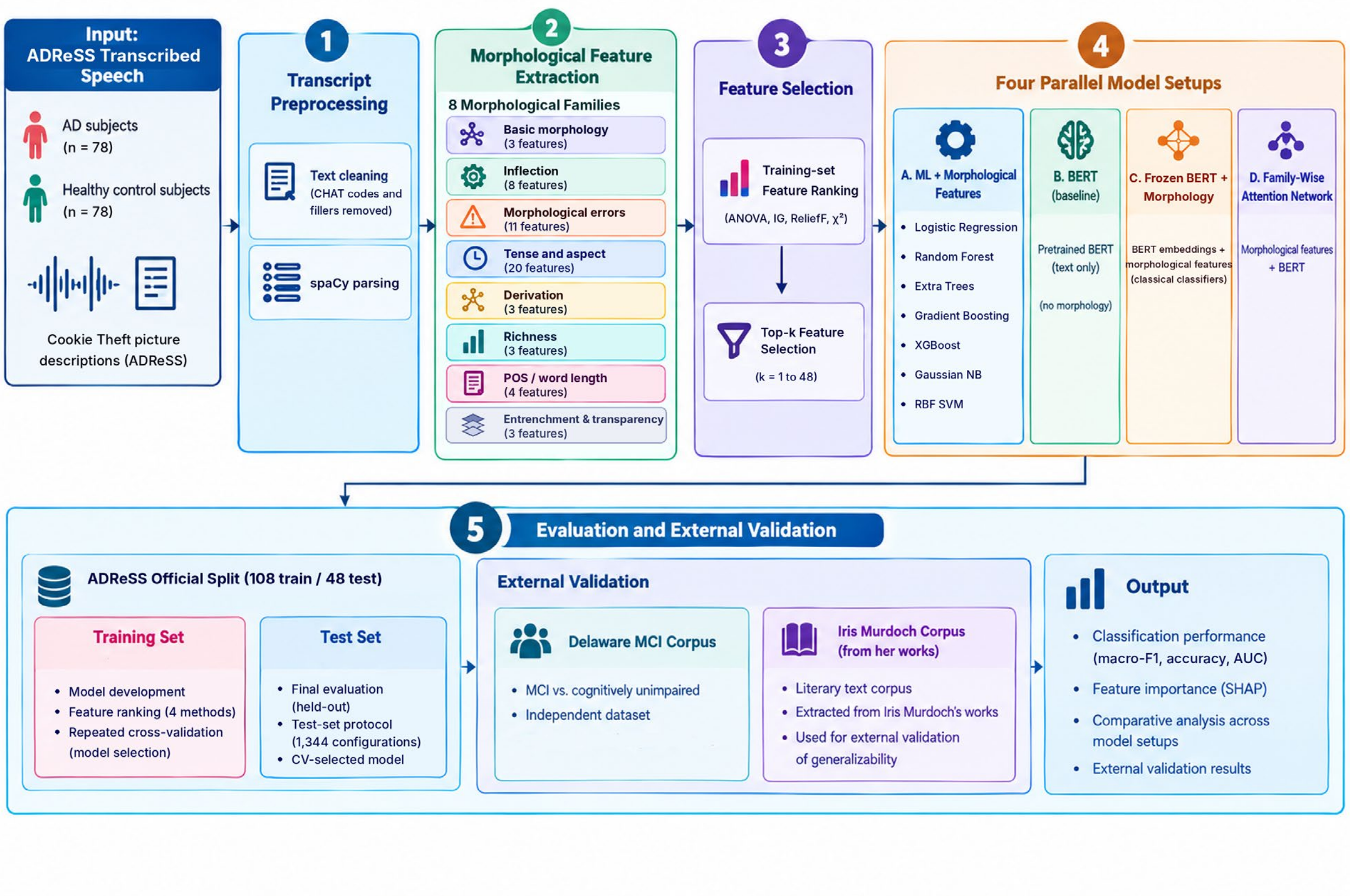

**Figure 2.** Rank-biserial effect sizes (95% bootstrap CI) for the 38 analyzed measures, grouped by family (extended feature set). Filled markers: $q < .05$ (Benjamini–Hochberg). Negative values indicate lower values in AD. * New theory-motivated measure.

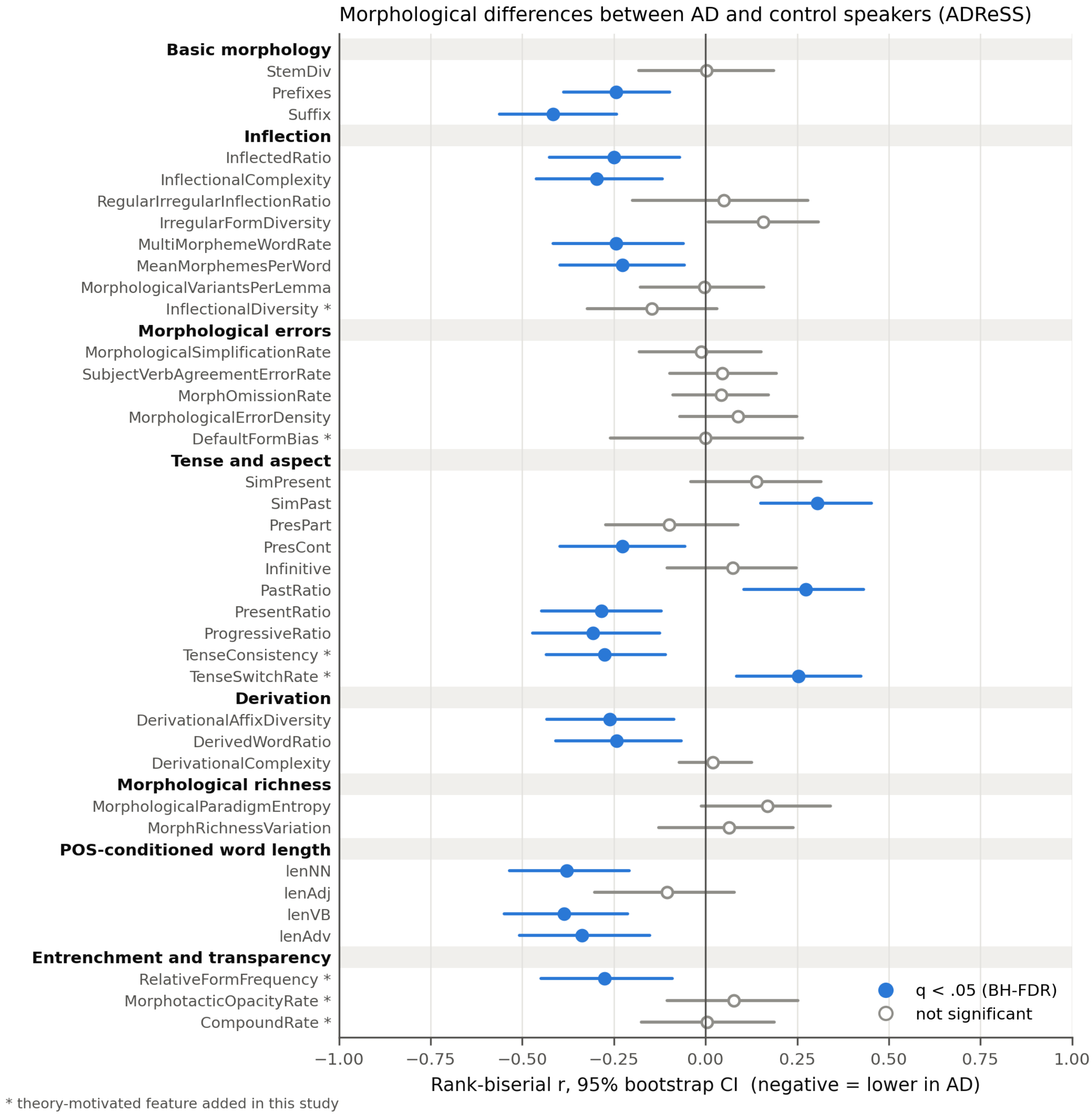

**Figure 3.** SHAP summary plots for the 48 test speakers (core feature set). For each ranking (ANOVA, information gain, ReliefF, $\chi^2$), a random forest was trained on the top 20 training-set features. Each point is a speaker; color shows the feature value; positive SHAP values push the prediction towards AD.

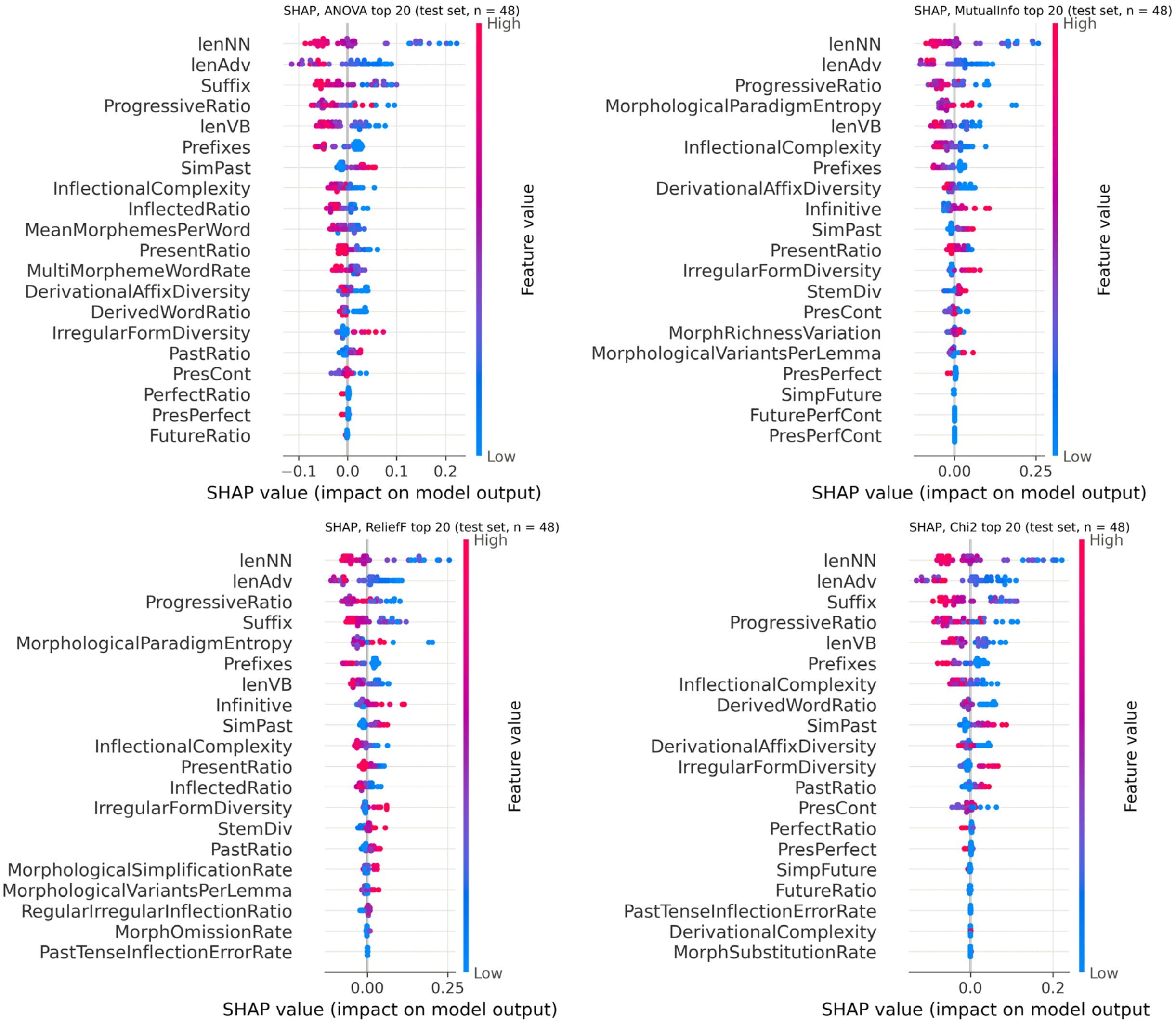

**Figure 4.** Feature rankings and their stability (training set, extended feature set). (a) Rank of the 20 top consensus measures under five ranking approaches: ANOVA, mutual information, ReliefF, $\chi^2$, and L1-regularized logistic regression path (1 = highest rank). (b) Proportion of 100 stratified 80% training subsamples in which each measure entered the top 10. Bars show the mean across approaches; markers show each approach.

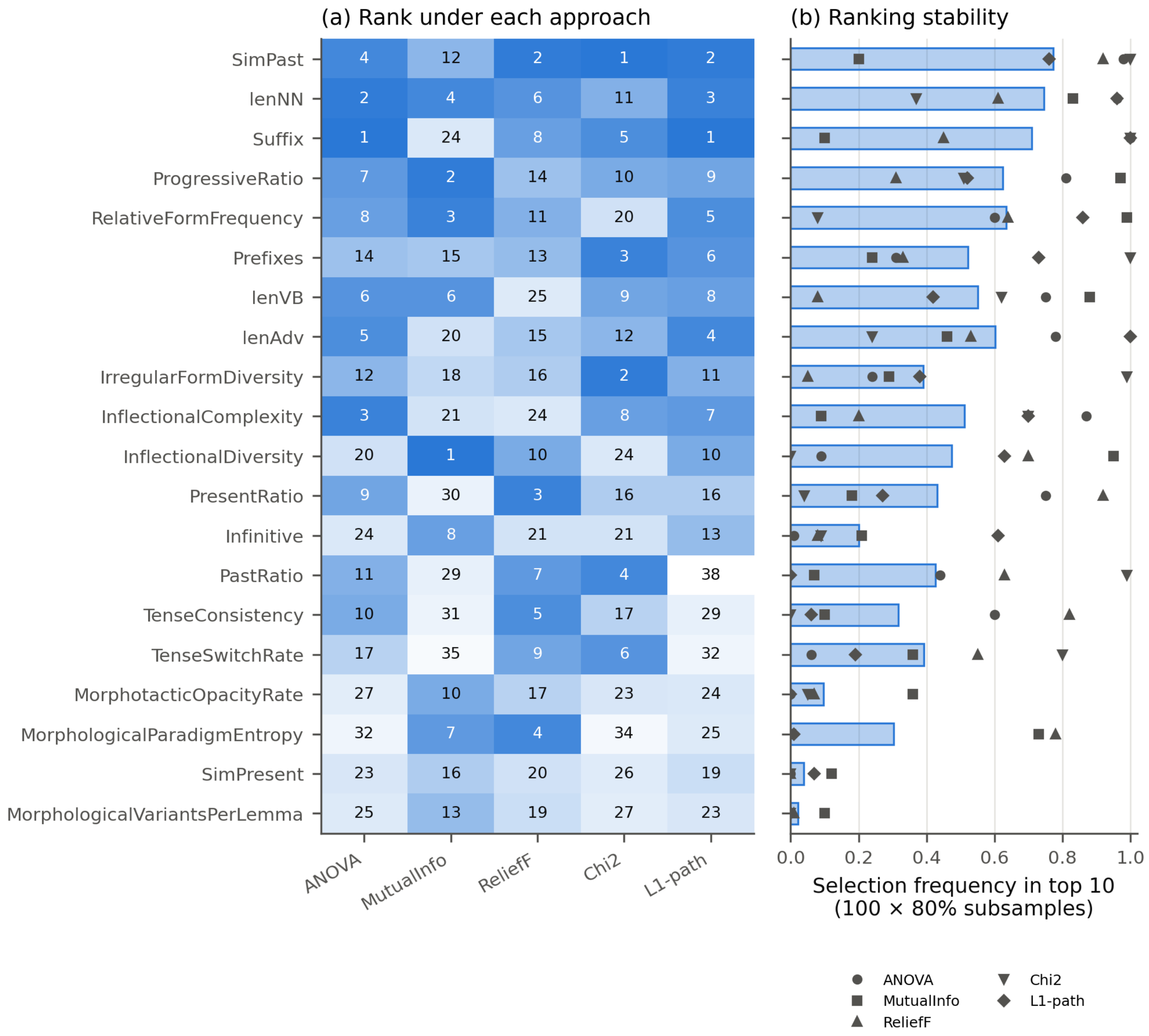

**Figure 5.** Family ablation on the training set (extended feature set; repeated 5-fold cross-validation). (a) Cross-validated AUC of models using a single family; the dashed line shows the model with all families. (b) AUC lost when a family is removed (leave-one-family-out) or jointly permuted (grouped permutation importance).

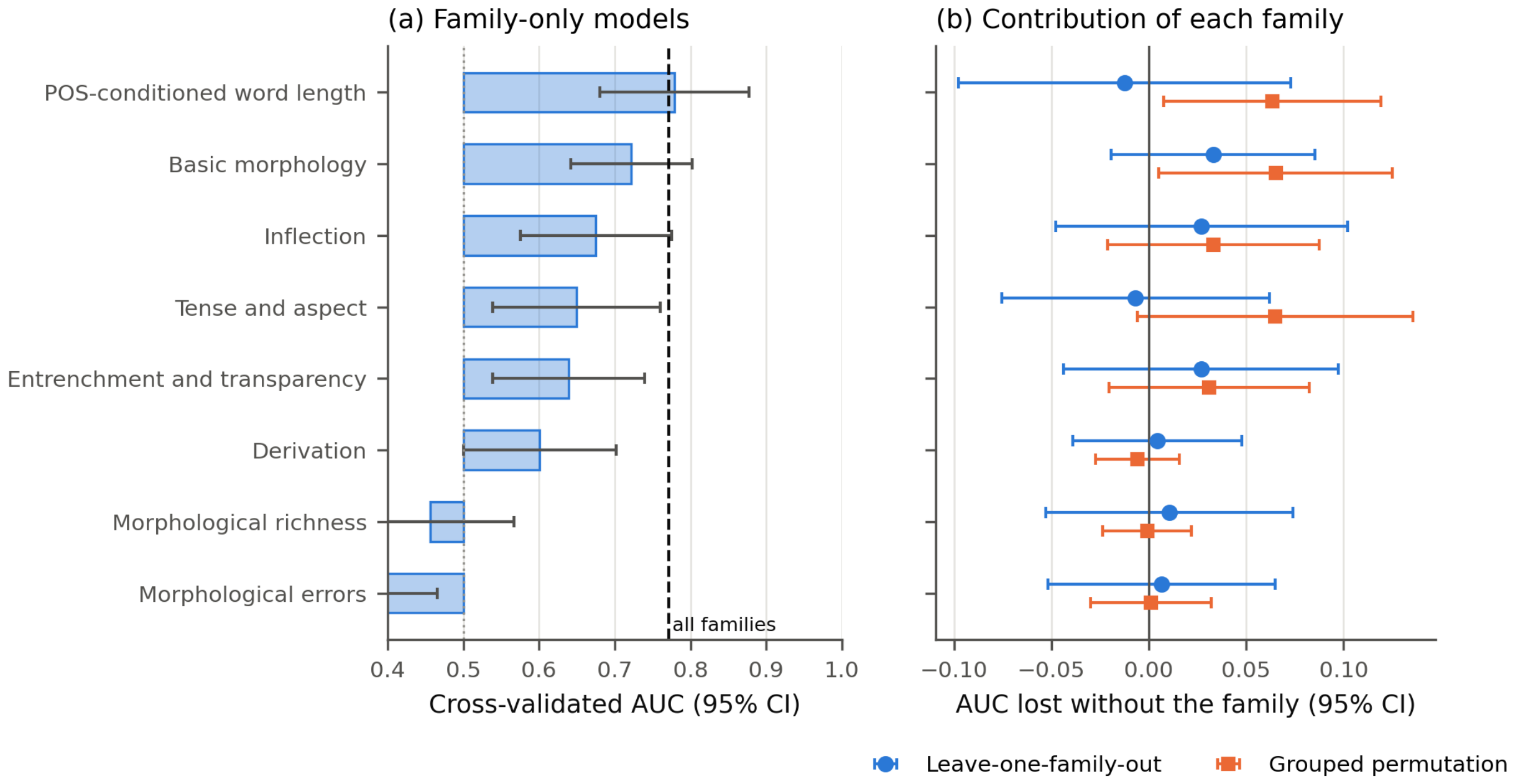

**Figure 6.** Macro-F1 on the ADReSS test set as a function of the number of selected features, for four ranking approaches and seven classifiers (core feature set). Features were ranked on the training set only; each model was trained on the training set and evaluated on the 48 test speakers. Markers show the best value of each classifier.

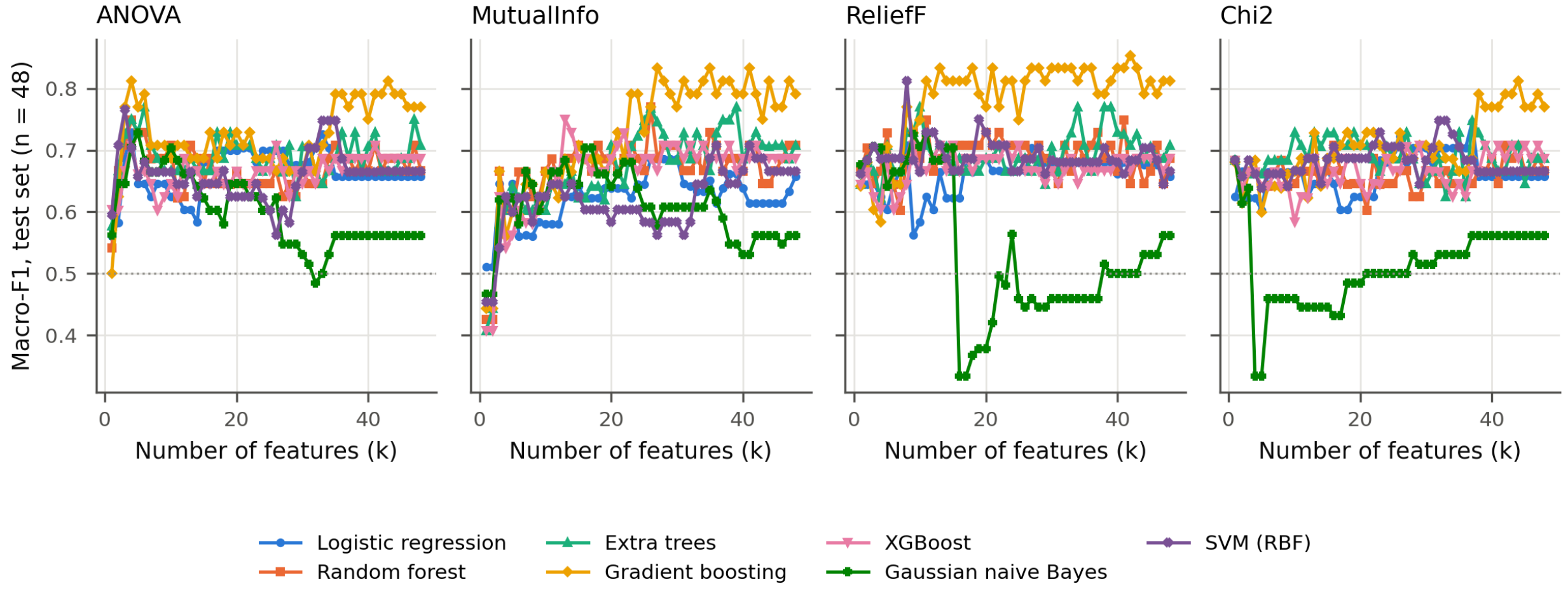

**Figure 7.** Confusion matrix of the best test-set configuration (gradient boosting with the 42 top-ranked ReliefF features, core feature set) on the 48 ADReSS test speakers. HC = healthy controls; AD = Alzheimer's disease.

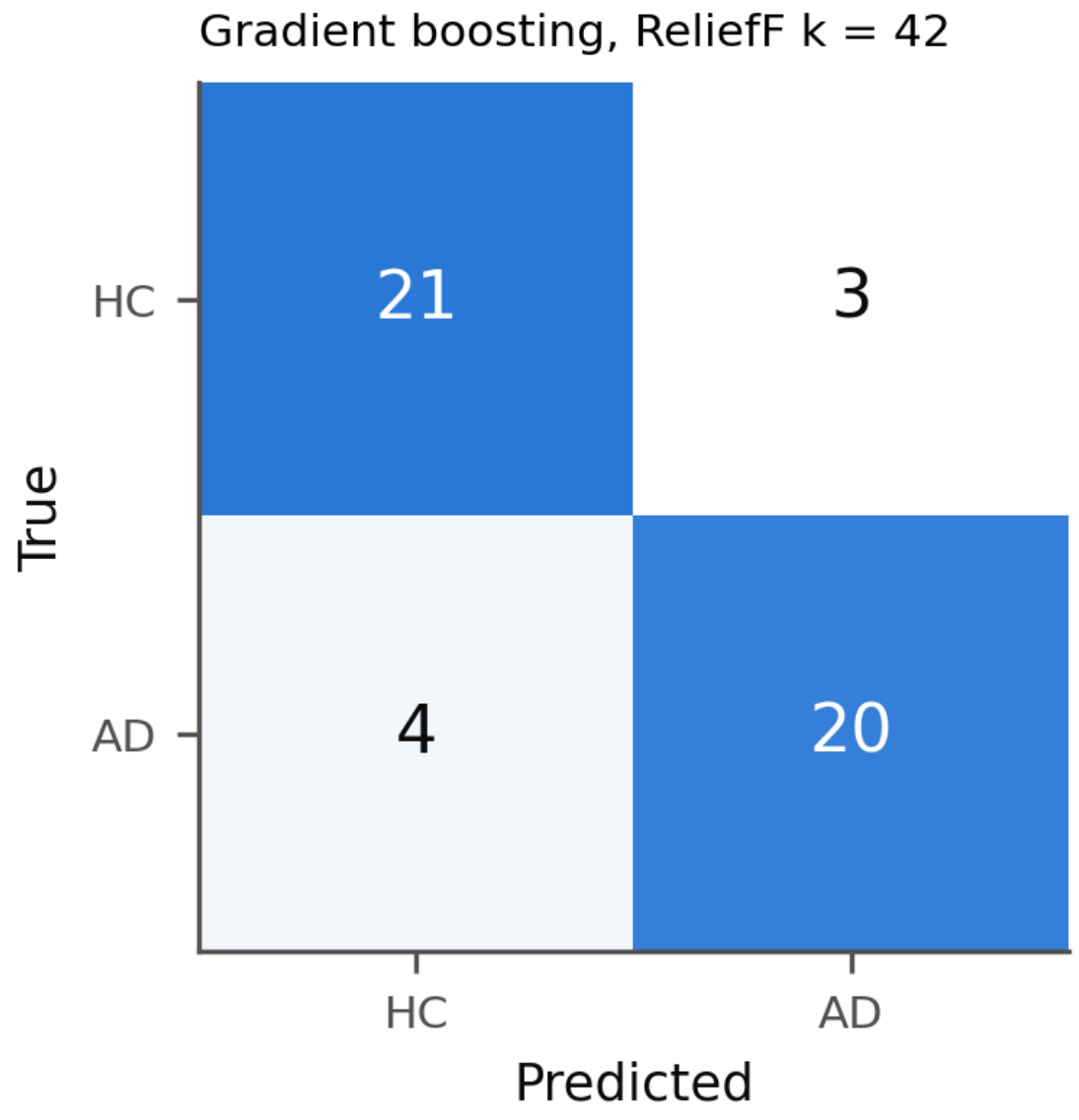

**Figure 8.** Mean cross-validated AUC on the training set as a function of the number of selected measures, for each ranking approach and classifier (extended feature set; rankings recomputed within each fold).

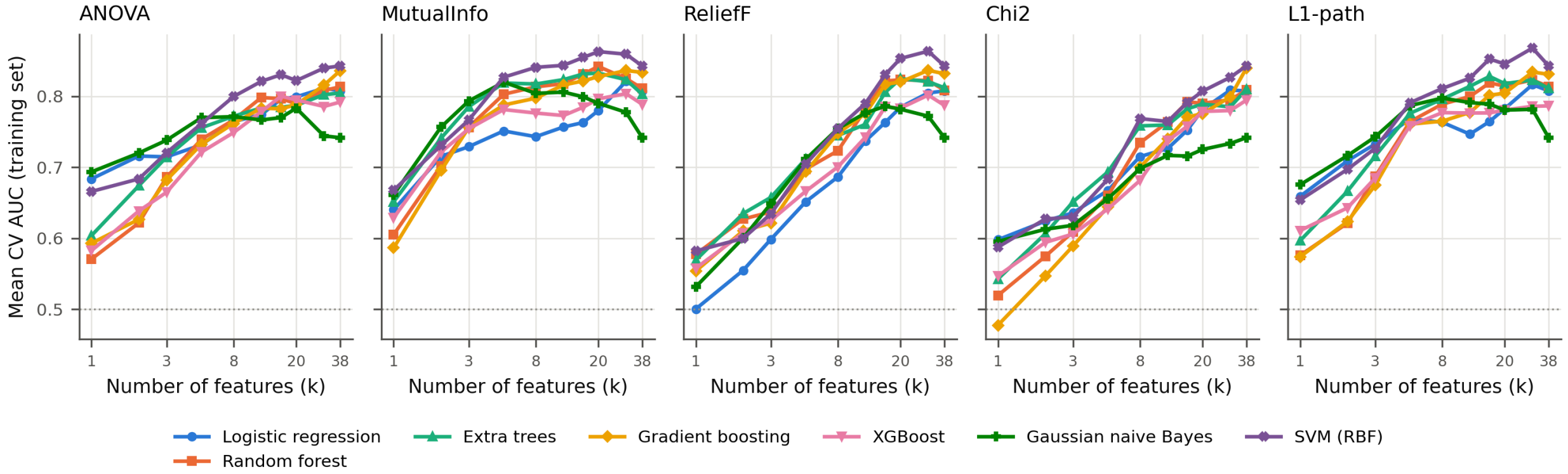

**Figure 9.** Mean cross-validated macro-F1 on the training set as a function of the number of selected measures, for each ranking approach and classifier (extended feature set; rankings recomputed within each fold).

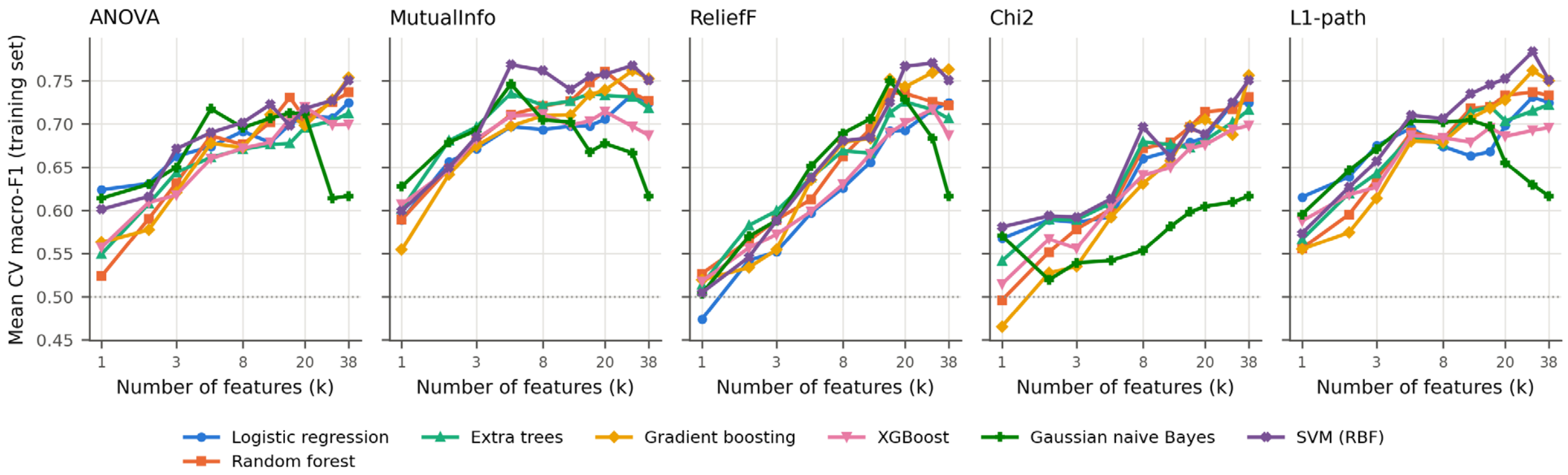

**Figure 10.** Mean attention weight assigned to each morphological family by the BERT + morphological-family attention model (48 test speakers; core feature set).

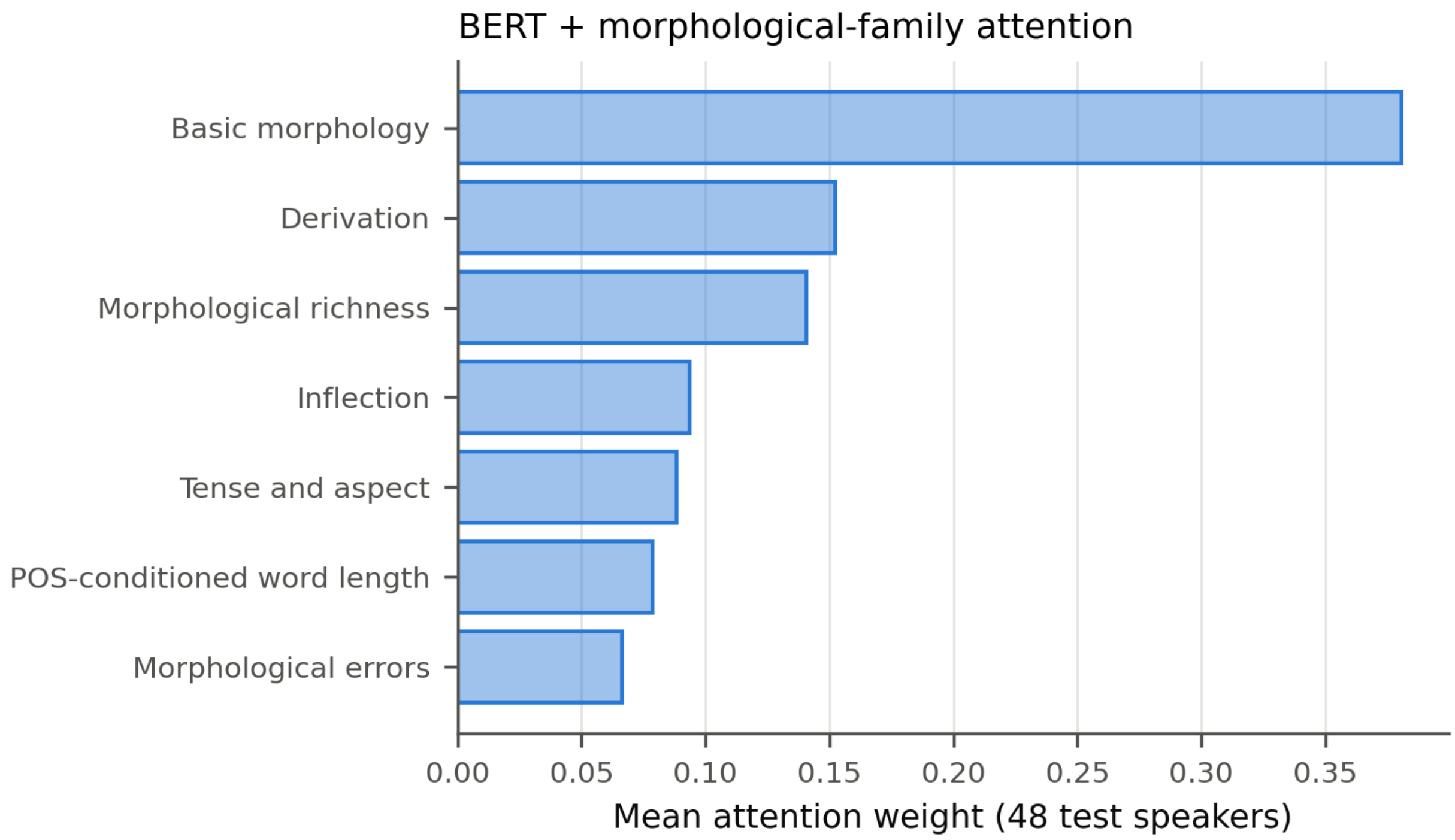

**Figure 11.** Log odds ratios per SD for the RQ1 measures before (open symbols) and after (filled symbols, 95% CI) adjustment for age, sex, transcript length, lexical diversity, and lexical frequency (extended feature set).

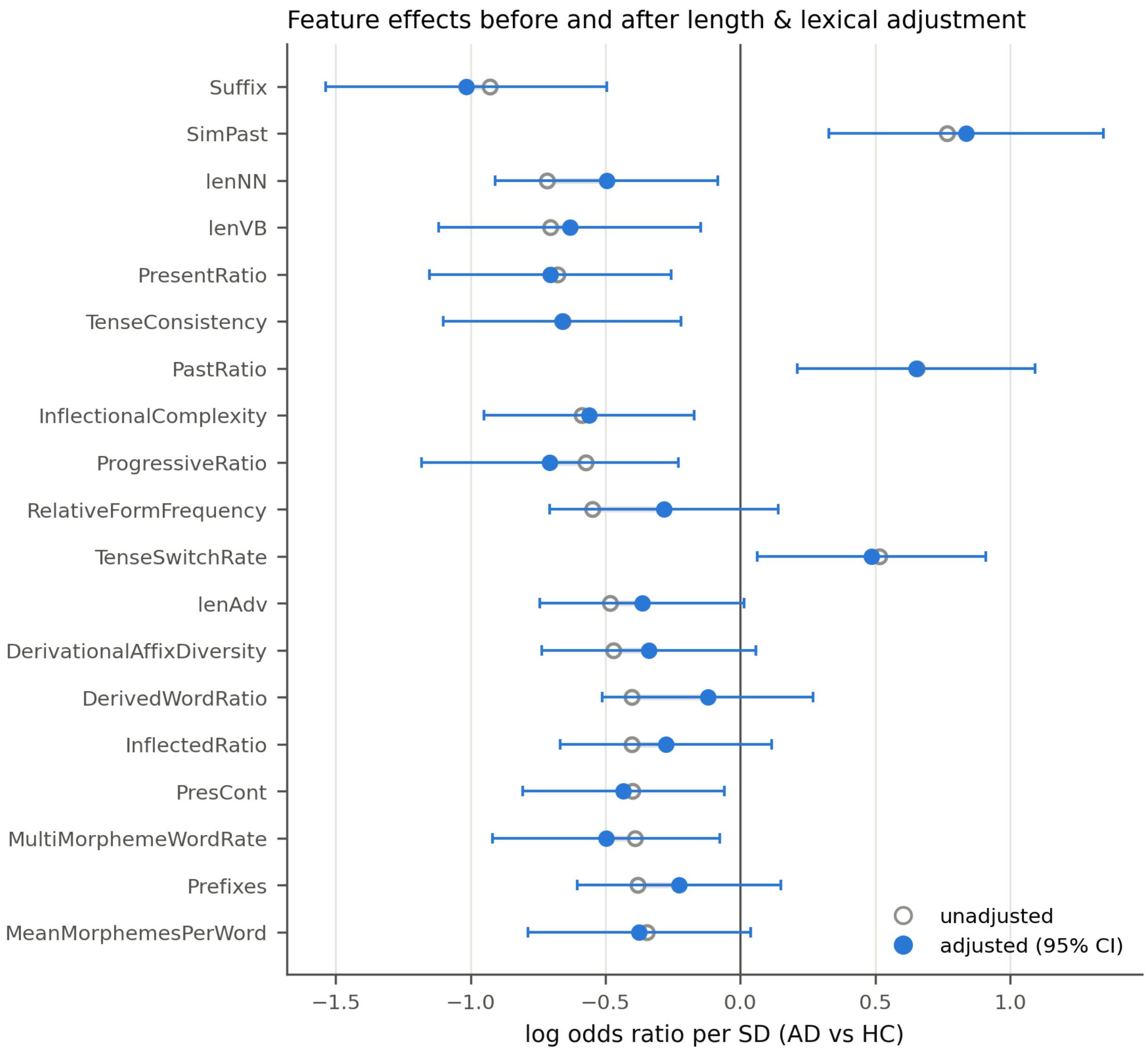

**Figure 12.** ROC curves of the three pre-specified logistic regression models on the held-out ADReSS test set (n = 48; extended feature set): controls only (age, sex, length, MATTR, lexical frequency; AUC 0.70), morphology only (AUC 0.69), and controls plus morphology (AUC 0.73). Dashed line: chance.

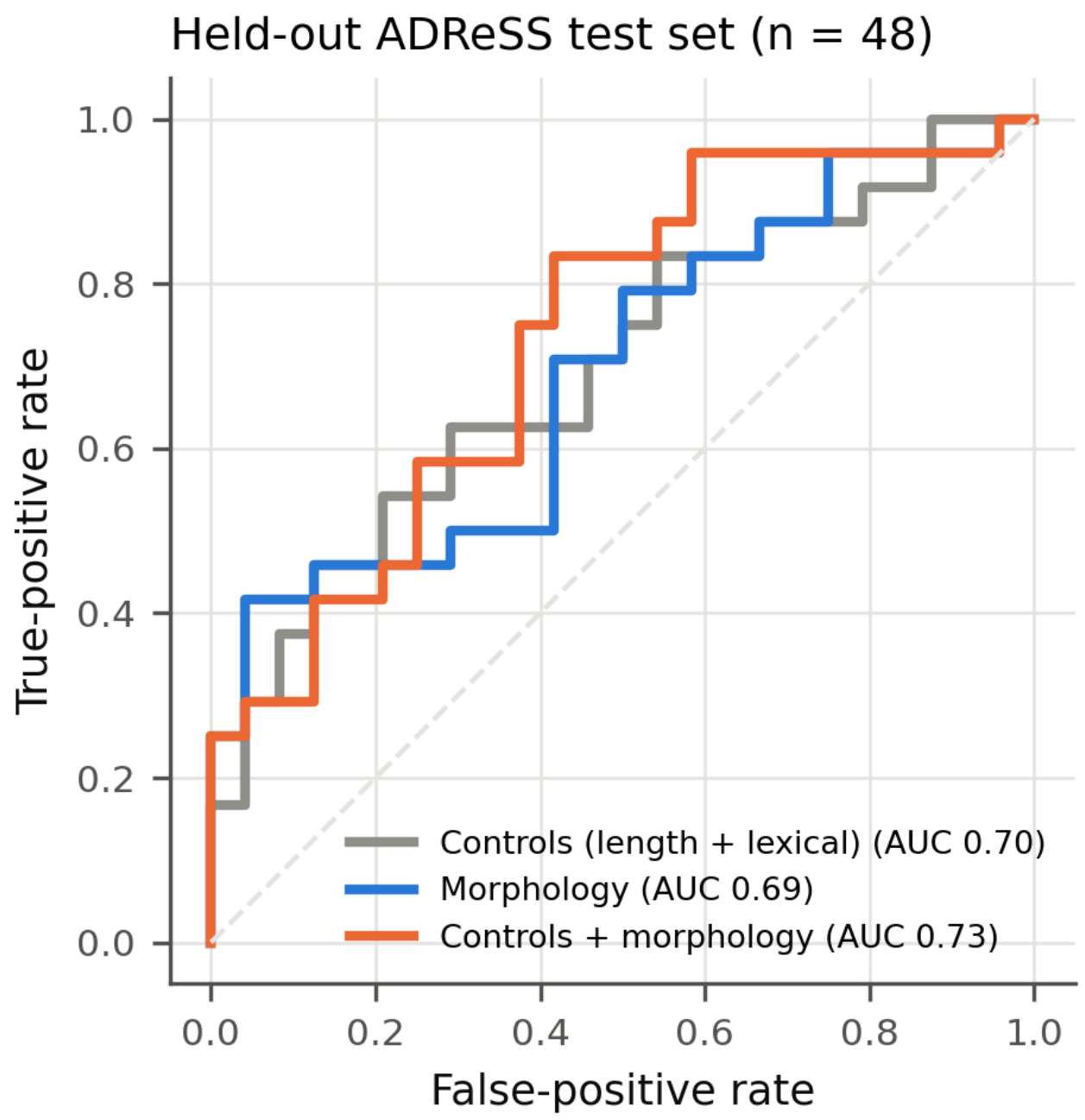

**Figure 13.** Effect sizes computed from full transcripts versus from the first 50 words of each transcript (extended feature set). Points near the diagonal indicate effects that do not depend on transcript length.

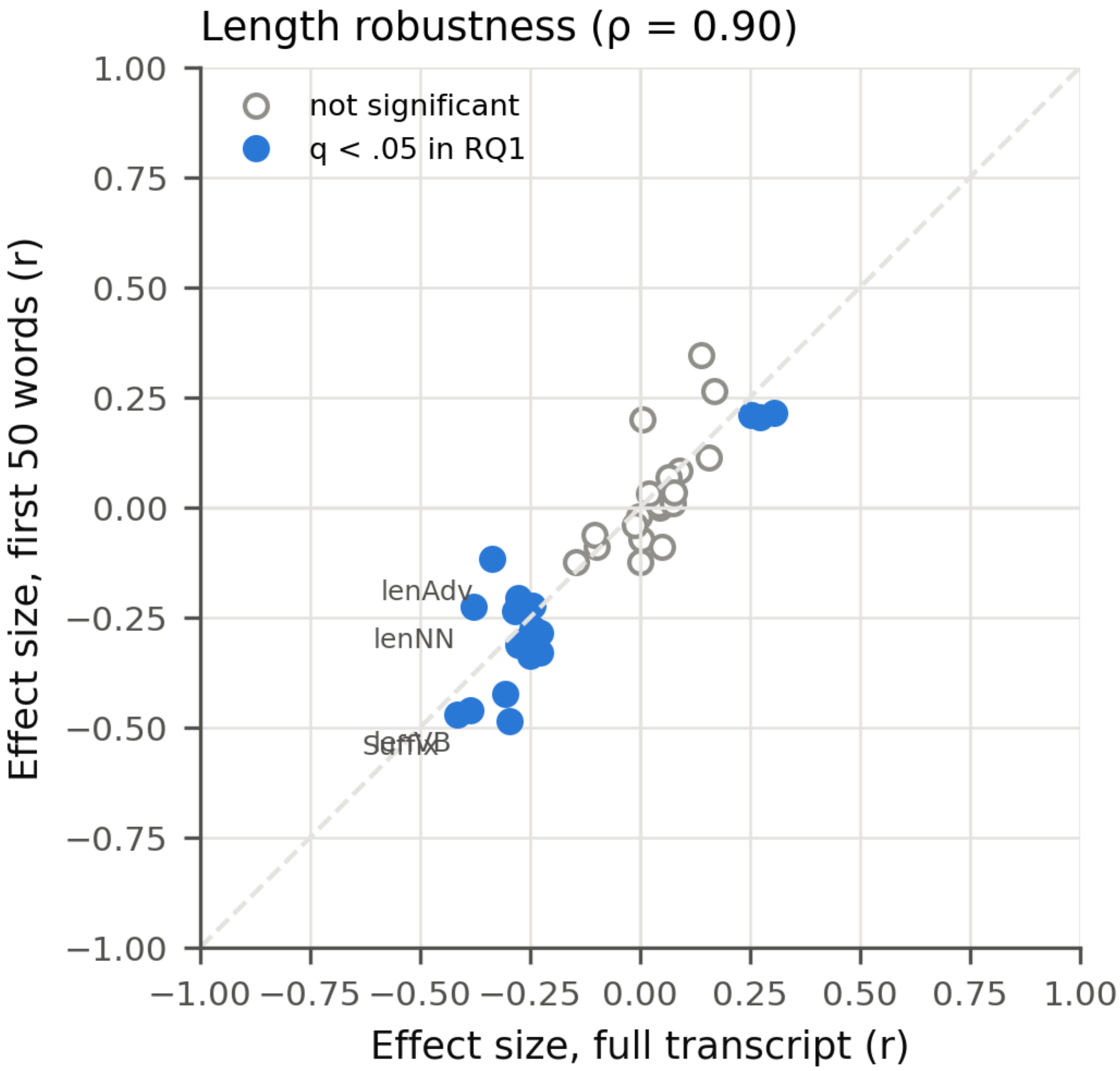

**Figure 14.** Exploratory replication of the direction of the 19 RQ1 effects in ADReSS (AD vs. controls), Delaware (MCI vs. controls), and the Iris Murdoch novels (late vs. early).

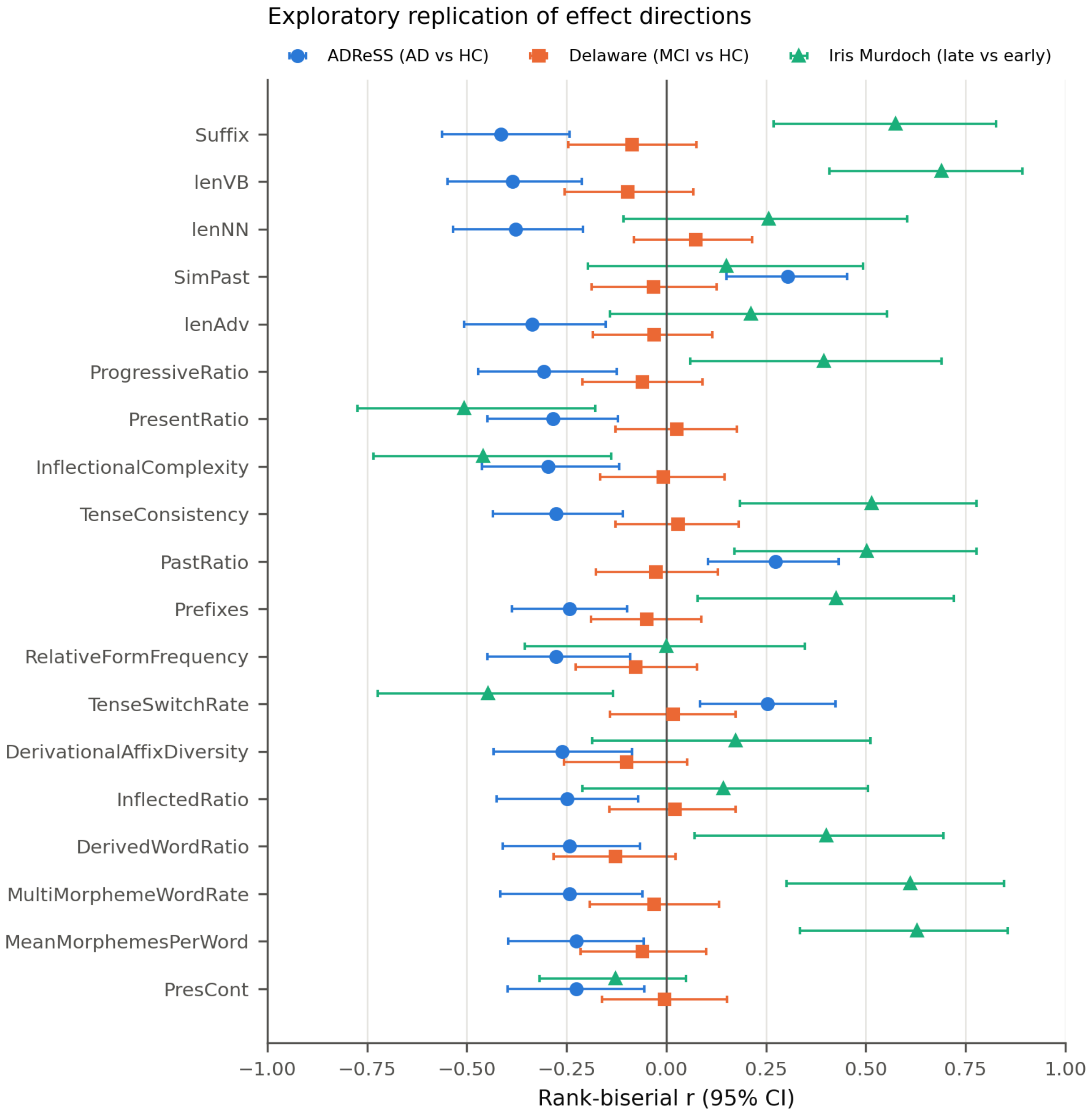

**Figure 15.** Transfer of ADReSS-trained models (core feature set) without adaptation: macro-F1 on Delaware (left column; MCI vs. controls, 291 sessions) and on the Iris Murdoch passages (right column; late vs. early, 44 passages) as a function of the number of selected features. Rows: ranking approach (ANOVA, information gain, ReliefF, $\chi^2$). Lines: classifiers. Dashed line: chance (0.5).

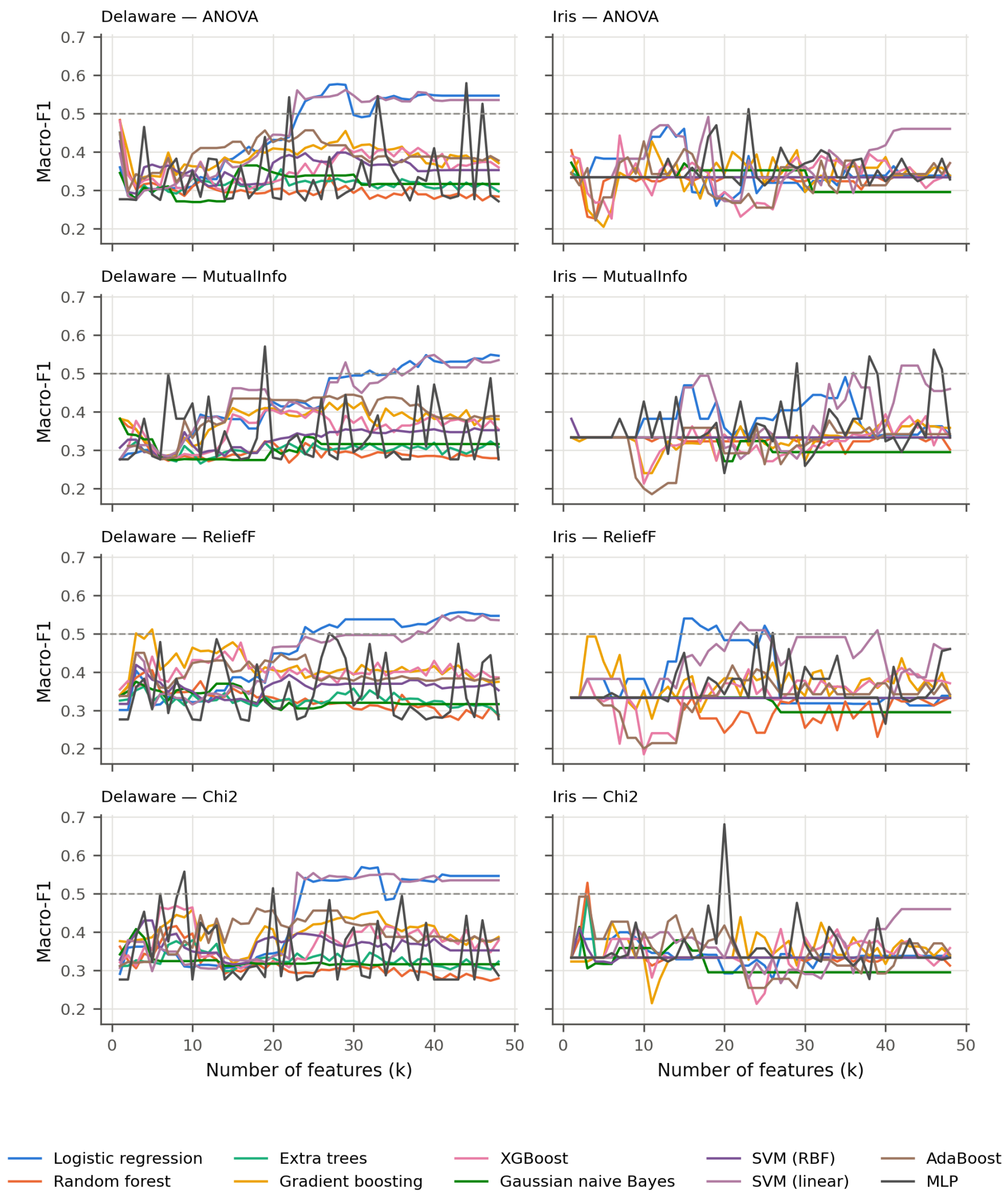

# Supplemental Material

All analyses use the participant-only ADReSS transcripts (78 AD, 78 controls; 108 training and 48 test speakers) and the same pipeline, software versions, and random seeds as the main text. Tables S1–S5 report additional classification analyses, Tables S6–S9 the validation of the automatic measures, and Tables S10–S11 the transfer analyses.

**Table S1**

*Ensembles of the best classifiers on the ADReSS test set (48 speakers). Each classifier enters with its own best ranking and number of features.*

| Model | CV macro-F1 (training) | Test macro-F1 [95% CI] | Test AUC |
|---|---|---|---|
| ***A. Members chosen by cross-validation on the training set*** | | | |
| Member: Extra trees (16 features) | 0.768 | 0.646 | 0.700 |
| Member: Gaussian naive Bayes (16 features) | 0.729 | 0.684 | 0.727 |
| Member: Gradient boosting (38 features) | 0.703 | 0.729 | 0.788 |
| Member: Logistic regression (30 features) | 0.721 | 0.622 | 0.688 |
| Member: Random forest (20 features) | 0.806 | 0.646 | 0.720 |
| Member: SVM (RBF) (30 features) | 0.814 | 0.624 | 0.747 |
| Member: XGBoost (20 features) | 0.694 | 0.583 | 0.663 |
| Ensemble: Hard voting | 0.777 | 0.624 [0.479, 0.750] | 0.725 |
| Ensemble: Soft voting | 0.740 | 0.664 [0.519, 0.791] | 0.753 |
| Ensemble: Weighted soft voting | 0.740 | 0.686 [0.541, 0.812] | 0.748 |
| Ensemble: Rank averaging | 0.768 | 0.687 [0.558, 0.812] | 0.744 |
| Ensemble: Stacking (logistic regression) | 0.768 | 0.667 [0.521, 0.792] | 0.757 |
| Ensemble: Greedy ensemble selection | 0.814 | 0.624 [0.473, 0.750] | 0.747 |
| Ensemble: Top-3 soft voting | 0.768 | 0.646 [0.499, 0.771] | 0.734 |
| ***B. Members chosen on the test set (upper bound)*** | | | |
| Member: Extra trees (6 features) | 0.675 | 0.771 | 0.804 |
| Member: Gaussian naive Bayes (5 features) | 0.749 | 0.728 | 0.773 |
| Member: Gradient boosting (42 features) | 0.684 | 0.854 | 0.847 |
| Member: Logistic regression (37 features) | 0.639 | 0.746 | 0.762 |

| Model | CV macro-F1 (training) | Test macro-F1 [95% CI] | Test AUC |
|---|---|---|---|
| Member: Random forest (26 features) | 0.722 | 0.770 | 0.792 |
| Member: SVM (RBF) (8 features) | 0.756 | 0.812 | 0.818 |
| Member: XGBoost (13 features) | 0.676 | 0.750 | 0.724 |
| Ensemble: Hard voting | 0.693 | 0.792 [0.666, 0.896] | 0.855 |
| Ensemble: Soft voting | 0.702 | 0.771 [0.646, 0.875] | 0.835 |
| Ensemble: Weighted soft voting | 0.712 | 0.792 [0.666, 0.896] | 0.832 |
| Ensemble: Rank averaging | 0.667 | 0.750 [0.622, 0.854] | 0.814 |
| Ensemble: Stacking (logistic regression) | 0.768 | 0.729 [0.596, 0.854] | 0.818 |
| Ensemble: Greedy ensemble selection | 0.756 | 0.812 [0.687, 0.917] | 0.818 |
| Ensemble: Top-3 soft voting | 0.730 | 0.729 [0.590, 0.853] | 0.816 |

Note. CV macro-F1 = out-of-fold macro-F1 on the 108 training speakers (slightly optimistic for stacking and greedy selection, whose combination is learned on the same predictions). Weights, the stacking meta-learner, and greedy selection were learned on training predictions only. Stacking coefficients, A: Extra trees: +1.27, Gaussian naive Bayes: +1.00, Gradient boosting: +0.50, Logistic regression: +1.19, Random forest: +1.12, SVM (RBF): +1.03, XGBoost: +0.09; greedy selection kept only the RBF SVM in both protocols; top-3 soft voting used, A: SVM (RBF), Random forest, Extra trees; B: SVM (RBF), Gaussian naive Bayes, Random forest. 95% CI: stratified bootstrap (2,000 resamples).

**Table S2**

*Deep-learning models on the ADReSS test set, single run with the default seed (seed 42).*

| **Model** | **Selection** | **Macro-F1 [95% CI]** | **AUC [95% CI]** | **Sens. / Spec.** |
|---|---|---|---|---|
| Fine-tuned BERT | validation split (early stopping) | 0.746 [0.614, 0.854] | 0.875 [0.767, 0.958] | 0.625 / 0.875 |
| Frozen BERT + morphology, XGBoost | chosen on validation split | 0.770 | 0.840 | 0.833 / 0.708 |
| BERT + family attention, final epoch | fixed (5 epochs) | 0.874 [0.768, 0.958] | 0.901 [0.799, 0.977] | 0.958 / 0.792 |
| BERT + morphology, gated fusion | checkpoint with best validation macro-F1 | 0.689 [0.547, 0.806] | 0.851 [0.726, 0.946] | 0.458 / 0.958 |
| Morphology-only MLP (no text) | randomized search, 5-fold CV | 0.646 [0.510, 0.770] | 0.646 [0.484, 0.797] | 0.625 / 0.667 |

Note. Gated fusion: fused representation = gate × [CLS] + (1 − gate) × projected morphology; mean gate (share of BERT) = 0.52. Morphology-only MLP: multilayer perceptron on the 48 core measures without text. Seed-to-seed variation is shown in Table S4.

**Table S3**

*Frozen BERT [CLS] embeddings plus the 48 core measures with eight classical classifiers.*

| Classifier | CV macro-F1 | Validation macro-F1 | Test macro-F1 [95% CI] | Test AUC |
|---|---|---|---|---|
| XGBoost | 0.798 | 0.768 | 0.770 [0.644, 0.875] | 0.840 |
| Gradient boosting | 0.836 | 0.718 | 0.791 [0.666, 0.896] | 0.812 |
| Logistic regression | 0.848 | 0.681 | 0.728 [0.596, 0.833] | 0.830 |
| Gaussian naive Bayes | 0.753 | 0.676 | 0.812 [0.700, 0.917] | 0.821 |
| MLP | 0.822 | 0.676 | 0.812 [0.687, 0.917] | 0.851 |
| Random forest | 0.801 | 0.633 | 0.833 [0.723, 0.917] | 0.878 |
| SVM | 0.835 | 0.633 | 0.750 [0.622, 0.855] | 0.830 |
| Extra trees | 0.824 | 0.590 | 0.833 [0.723, 0.917] | 0.880 |

Note. Hyperparameters tuned by randomized search with 5-fold cross-validation on the training split; the classifier reported in the main text (XGBoost) is the one with the best validation macro-F1. Test values of the other classifiers are shown for transparency and should not be used to select a model.

**Table S4**

*Seed robustness of the BERT-based models: test macro-F1 / AUC for five random seeds (same training, validation, and test speakers).*

| Seed | Fine-tuned BERT | BERT + family attention | BERT + morphology, gated fusion |
|---|---|---|---|
| 42 | 0.791 / 0.901 | 0.874 / 0.901 | 0.689 / 0.851 |
| 1042 | 0.790 / 0.901 | 0.854 / 0.918 | 0.726 / 0.858 |
| 2042 | 0.791 / 0.865 | 0.812 / 0.905 | 0.792 / 0.856 |
| 3042 | 0.832 / 0.889 | 0.625 / 0.903 | 0.704 / 0.773 |
| 4042 | 0.790 / 0.891 | 0.768 / 0.901 | 0.750 / 0.812 |
| *Summary* | | | |
| Mean ± SD | 0.799 ± 0.019 / 0.889 ± 0.015 | 0.787 ± 0.099 / 0.906 ± 0.007 | 0.732 ± 0.041 / 0.830 ± 0.037 |
| Range (macro-F1) | 0.790–0.832 | 0.625–0.874 | 0.689–0.792 |
| Seed ensemble | 0.833 [0.719, 0.937] / 0.913 | 0.832 [0.723, 0.937] / 0.924 | 0.706 [0.571, 0.832] / 0.839 |

Note. Only weight initialization and batch order change between seeds. Seed ensemble = mean predicted probability over the five seeds, with a 95% stratified bootstrap CI for macro-F1.

**Table S5**

*Paired differences in test macro-F1 between fusion models and fine-tuned BERT across the five seeds.*

| Comparison | Mean difference (SD) | Range | Seeds better |
|---|---|---|---|
| BERT + family attention (final epoch) − fine-tuned BERT | -0.012 (0.116) | -0.207 to +0.083 | 3 of 5 |
| BERT + morphology, gated fusion − fine-tuned BERT | -0.067 (0.051) | -0.128 to +0.000 | 1 of 5 |

Note. In AUC, the family-attention model was at least as high as fine-tuned BERT in all five seeds (0.906 ± 0.007 vs. 0.889 ± 0.015).

**Table S6**

*Automatic error detector compared with the manual gold standard (27 verified errors in 156 participant-only transcripts).*

| Measure | Subset | k/n | Value | 95% CI |
|---|---|---|---|---|
| Precision | All | 11/109 | 0.101 | [0.057, 0.172] |
| Precision | AD | 10/58 | 0.172 | [0.096, 0.289] |
| Precision | HC | 1/51 | 0.020 | [0.003, 0.103] |
| Precision | agreement commission | 2/32 | 0.062 | [0.017, 0.201] |
| Precision | agreement omission 3sg | 5/30 | 0.167 | [0.073, 0.336] |
| Precision | bare stem after aux | 0/2 | 0.000 | [0.000, 0.658] |
| Precision | overregularization | 0/1 | 0.000 | [0.000, 0.793] |
| Precision | participle for past | 1/6 | 0.167 | [0.030, 0.564] |
| Precision | past for participle | 1/1 | 1.000 | [0.207, 1.000] |
| Precision | plural omission | 0/9 | 0.000 | [0.000, 0.299] |
| Precision | plural with singular det | 2/4 | 0.500 | [0.150, 0.850] |
| Precision | pronoun case | 0/24 | 0.000 | [0.000, 0.138] |
| Recall | All | 10/27 | 0.370 | [0.215, 0.558] |
| Recall | Certain | 2/5 | 0.400 | [0.118, 0.769] |
| Recall | Probable | 4/12 | 0.333 | [0.138, 0.609] |
| Recall | Unsure | 4/10 | 0.400 | [0.168, 0.687] |

Note. Precision = share of automatic detections that match a verified error (same speaker, within two words); recall = share of verified errors detected. Recall subsets refer to the annotator's confidence. Wilson confidence intervals.

**Table S7**

*Manually verified morphological errors by group (78 AD, 78 HC).*

| Errors | AD: errors (speakers) | HC: errors (speakers) | AD per 1,000 words | HC per 1,000 words | Fisher p (speakers) | Mann–Whitney p (rates) |
|---|---|---|---|---|---|---|
| All verified | 20 (15) | 7 (7) | 2.63 | 0.80 | 0.106 | 0.061 |
| Certain + Probable | 13 (12) | 4 (4) | 1.71 | 0.46 | 0.062 | 0.033 |
| 3sg omission | 5 (5) | 0 (0) | 0.66 | 0.00 | 0.059 | 0.024 |

Note. Fisher's exact test compares the number of speakers with at least one error; the Mann–Whitney test compares per-speaker error rates.

**Table S8**

*Sensitivity of the error measures to the gold standard: automatic detector, detector restricted to confirmed detections, and manually verified errors.*

| Measure | Version | Mean HC | Mean AD | r [95% CI] | Adjusted OR per SD [95% CI] | p (adj.) |
|---|---|---|---|---|---|---|
| ThirdPersonSOmissionRate | Detector | 0.018 | 0.032 | 0.093 [-0.02, 0.21] | 1.64 [1.10, 2.44] | 0.016 |
| ThirdPersonSOmissionRate | Detector, confirmed hits only | 0.000 | 0.008 | 0.052 [0.01, 0.10] | | n.e. |
| Errors per 100 words | Detector | 0.521 | 0.664 | 0.089 [-0.08, 0.26] | 1.45 [0.99, 2.11] | 0.055 |
| Errors per 100 words | Detector, confirmed hits only | 0.009 | 0.123 | 0.065 [0.00, 0.13] | 2.31 [0.67, 7.92] | 0.183 |
| Errors per 100 words | Manual verified errors | 0.079 | 0.254 | 0.105 [-0.00, 0.21] | 1.38 [0.83, 2.30] | 0.211 |

Note. Adjusted odds ratios from logistic regression with age, sex, log length, MATTR, and lexical frequency. n.e. = not estimable (too few non-zero values). The detector-based third-person-singular effect largely disappears when only confirmed detections are counted, because most detections in this category were false positives.

**Table S9**

*Accuracy of the automatic tense/aspect labels and morpheme counts on blinded random samples (balanced by group).*

| Subset | Correct/items | Accuracy | 95% CI |
|---|---|---|---|
| *Tense/aspect labels (162 verb groups)* | | | |
| All | 150/162 | 0.926 | [0.875, 0.957] |
| AD | 78/83 | 0.940 | [0.867, 0.974] |
| HC | 72/79 | 0.911 | [0.828, 0.956] |
| Bare -ing participle | 19/25 | 0.760 | |
| Present progressive | 44/44 | 1.000 | |
| Simple present | 43/46 | 0.935 | |
| Simple past | 18/20 | 0.900 | |
| Infinitive | 13/13 | 1.000 | |
| Modal | 8/8 | 1.000 | |
| Other categories (past progressive, bare past participle, present perfect, simple future) | 5/6 | 0.833 | |
| *Morpheme counts (100 words; 97 scorable)* | | | |
| All | 87/97 | 0.897 | [0.821, 0.943] |
| AD | 46/50 | 0.920 | [0.812, 0.968] |
| HC | 41/47 | 0.872 | [0.748, 0.940] |

Note. Category rows refer to the category assigned by the pipeline. AD vs. HC: Fisher p = .56 (tense/aspect) and p = .52 (morpheme counts). Three word fragments were excluded from the morpheme sample. Systematic errors: progressive forms after a noun subject with contracted is (e.g., the girl's reaching) labeled as bare participles (5 of 6 bare-participle errors; 4 in HC), and derivational affixes not recognized (-ly, -ery, -ion) or wrongly assigned (cookie, valance). Items were labeled by the author and reviewed by an AI model; disagreements were adjudicated against explicit rules.

**Table S10**

*Transfer without adaptation: threshold-free AUC, macro-F1 at the ADReSS threshold, and share of samples predicted as impaired.*

| Model (trained on all 156 ADReSS speakers) | AUC [95% CI] | Macro-F1 | Share predicted impaired |
|---|---|---|---|
| *Delaware (291 samples, 111 MCI)* | | | |
| Extra trees | 0.512 [0.443, 0.581] | 0.273 | 0.98 |
| Gaussian naive Bayes | 0.506 [0.443, 0.574] | 0.318 | 0.92 |
| Gradient boosting | 0.452 [0.382, 0.520] | 0.391 | 0.82 |
| Logistic regression | 0.558 [0.490, 0.624] | 0.557 | 0.35 |
| Random forest | 0.498 [0.429, 0.566] | 0.300 | 0.96 |
| SVM (RBF) | 0.493 [0.426, 0.566] | 0.296 | 0.95 |
| XGBoost | 0.444 [0.377, 0.513] | 0.342 | 0.90 |
| Soft-voting ensemble | 0.500 [0.431, 0.569] | 0.317 | 0.95 |
| *Iris (44 samples, 22 late passages)* | | | |
| Extra trees | 0.167 [0.058, 0.310] | 0.323 | 0.98 |
| Gaussian naive Bayes | 0.523 [0.500, 0.568] | 0.333 | 1.00 |
| Gradient boosting | 0.281 [0.138, 0.453] | 0.407 | 0.30 |
| Logistic regression | 0.473 [0.285, 0.645] | 0.491 | 0.75 |
| Random forest | 0.209 [0.091, 0.355] | 0.281 | 0.73 |
| SVM (RBF) | 0.205 [0.081, 0.360] | 0.333 | 1.00 |
| XGBoost | 0.260 [0.128, 0.409] | 0.338 | 0.57 |
| Soft-voting ensemble | 0.260 [0.126, 0.426] | 0.302 | 0.93 |

Note. Each classifier uses its cross-validated ranking and number of features (no selection on the target data); the soft-voting ensemble averages their probabilities. AUC < 0.5 means that the ADReSS-trained scores rank the target groups in the opposite direction.

**Table S11**

*Transfer of BERT-based models trained on all 156 ADReSS speakers (macro-F1, no adaptation).*

| Model | Delaware (MCI vs. controls) | Murdoch (late vs. early) |
|---|---|---|
| BERT + family attention (final epoch) | 0.463 | 0.302 |
| Fine-tuned BERT | 0.558 | 0.500 |

Note. Default seed; fine-tuned BERT used the number of epochs chosen by early stopping on ADReSS.

**Figure S1.** Test macro-F1 of the member classifiers (gray) and of the ensembles (blue) under the two protocols of Table S1.

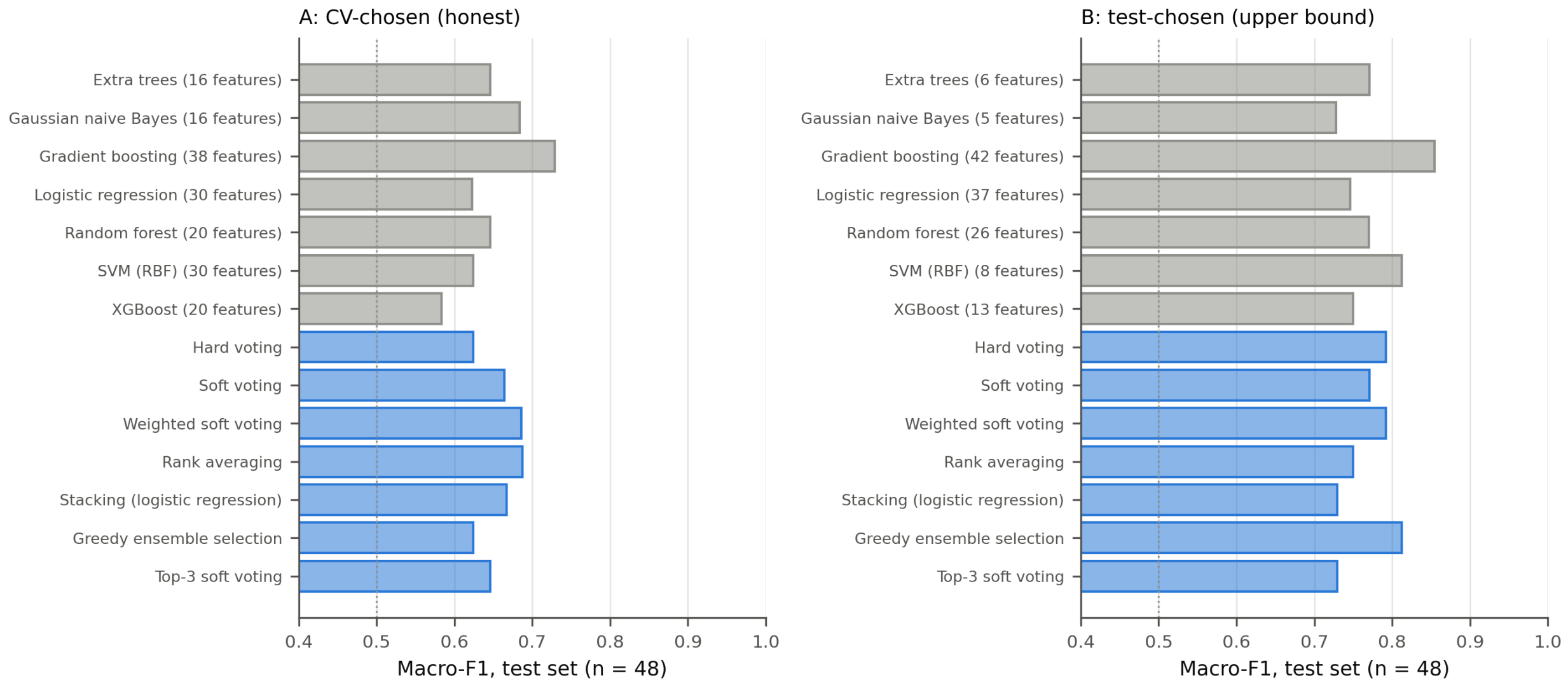